\UseRawInputEncoding
\documentclass[10pt,twocolumn]{article} 
\pdfoutput=1
\usepackage{simpleConference}
\usepackage{times}
\usepackage{graphicx}
\usepackage{amsmath}
\usepackage{amssymb}
\usepackage{booktabs}
\usepackage{algorithm}
\usepackage{algorithmic}
\usepackage{makecell}
\usepackage{cuted}
\usepackage{float}
\usepackage{capt-of}
\usepackage{hyperref}
\hypersetup{pagebackref,breaklinks,colorlinks}

\usepackage[capitalize]{cleveref}
\crefname{section}{Sec.}{Secs.}
\Crefname{section}{Section}{Sections}
\Crefname{table}{Table}{Tables}
\crefname{table}{Tab.}{Tabs.}

\begin{document}

\title{Few-shot Image Generation via Masked Discrimination}

\author{JingYuan Zhu \\
Tsinghua University, China \\
jy-zhu20@mails.tsinghua.edu.cn\\
\and
Huimin Ma \\
University of Science and Technology Beijing, China \\
mhmpub@ustb.edu.cn  \\
\and
Jiansheng Chen \\
University of Science and Technology Beijing, China \\
jschen@ustb.edu.cn  \\
\and
Jian Yuan  \\
Tsinghua University, China \\
jyuan@tsinghua.edu.cn  \\
}

\maketitle
\thispagestyle{empty}

\begin{strip}
\centering
\includegraphics[width=1.0\linewidth]{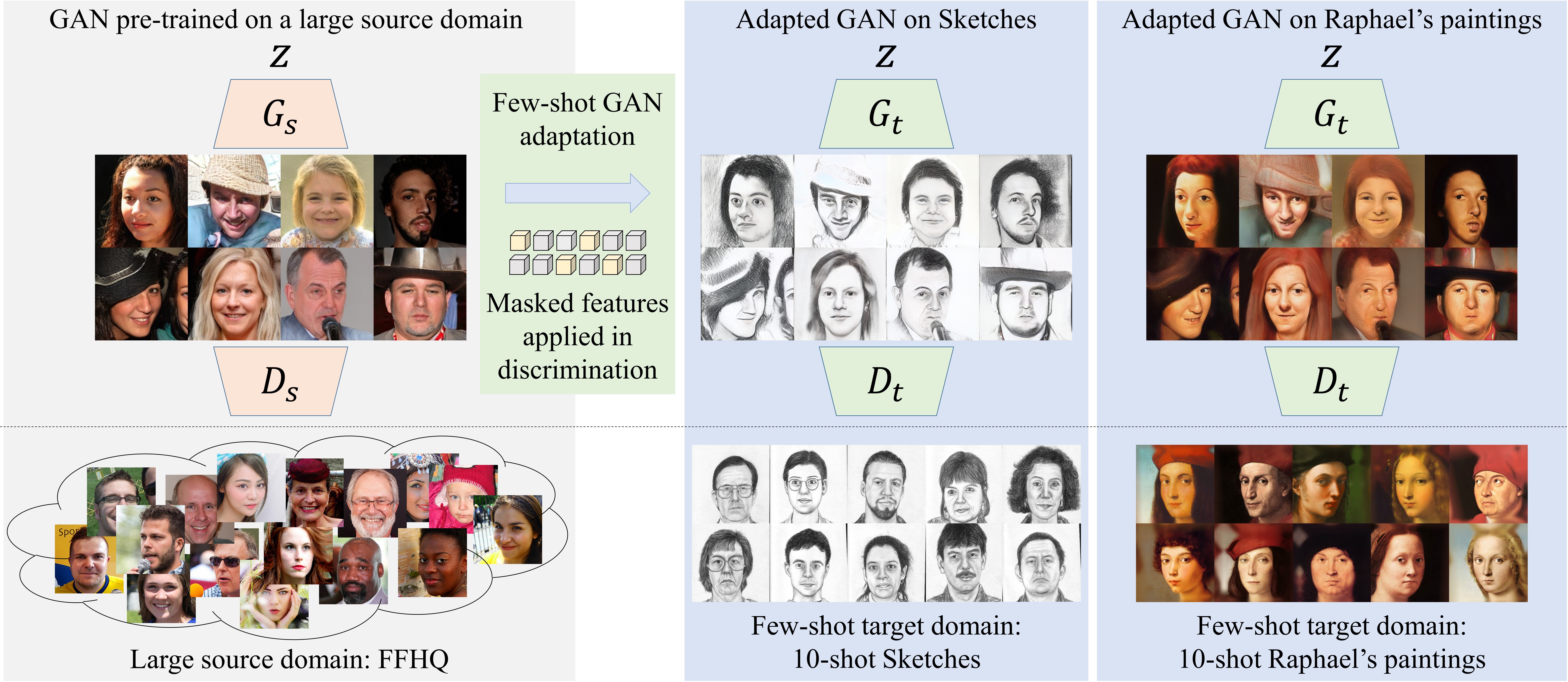}
\captionof{figure}{Based on a GAN model ($G_s + D_s$)  pre-trained on a large source domain (e.g., FFHQ), we propose to adapt it to target domains with limited data (e.g., 10 images) via masked discrimination (adapted GANs: $G_t + D_t$). Our approach can adapt the pre-trained source model to target domains naturally and maintain generation diversity.
\label{begin2}}
\end{strip}

\begin{abstract}
   Few-shot image generation aims to generate images of high quality and great diversity with limited data. However, it is difficult for modern GANs to avoid overfitting when trained on only a few images. The discriminator can easily remember all the training samples and guide the generator to replicate them, leading to severe diversity degradation. Several methods have been proposed to relieve overfitting by adapting GANs pre-trained on large source domains to target domains using limited real samples. This work presents a novel approach to realize few-shot GAN adaptation via masked discrimination. Random masks are applied to features extracted by the discriminator from input images. We aim to encourage the discriminator to judge various images which share partially common features with training samples as realistic. Correspondingly, the generator is guided to generate diverse images instead of replicating training samples. In addition, we employ a cross-domain consistency loss for the discriminator to keep relative distances between generated samples in its feature space. It strengthens global image discrimination and guides adapted GANs to preserve more information learned from source domains for higher image quality. The effectiveness of our approach is demonstrated both qualitatively and quantitatively with higher quality and greater diversity on a series of few-shot image generation tasks than prior methods.
\end{abstract}

%%%%%%%%% BODY TEXT
\section{Introduction}
    Modern generative adversarial networks (GANs)  \cite{NIPS2014_5ca3e9b1} containing abundant parameters have achieved great success in image generation with large amounts of training data. For example, BigGAN \cite{DBLP:conf/iclr/BrockDS19} makes use of more than 100M parameters to achieve significant improvement on ImageNet. StyleGAN \cite{Karras_2019_CVPR} is trained on 70,000 images from Flicker-Faces-HQ (FFHQ) \cite{Karras_2019_CVPR} with a total of 26.2M trainable parameters in its generator. However, GANs easily overfit and tend to replicate training data instead of generating diverse samples when trained on datasets containing fewer samples \cite{replicate2021}. Unfortunately, only a few samples can be obtained in some cases like famous artists' paintings.  

    For few-shot image generation tasks, the complex discriminator can remember all the training samples and excessively guide the generator to replicate them, leading to overfitting and serious diversity degradation. Several prior works \cite{ewc, mo2020freeze, noguchi2019image, wang2018transferring, wang2020minegan} have been proposed to adapt GANs pre-trained on related large source domains to target domains with limited data. They aim to preserve the diversity of adapted GANs by keeping information learned from source domains. Data augmentation approaches \cite{ada, zhao2020differentiable} also play positive roles in few-shot image generation. However, these approaches mostly need hundreds of training samples from target domains to produce high-quality results. When trained with extremely few samples (e.g., $\leq 10$ images), they cannot avoid overfitting or replicating training samples, resulting in poor generation results.

    Recent works \cite{ojha2021few-shot-gan,zhao2022closer} propose to build a one-to-one correspondence between the source and target domains, leading to greater diversity. A cross-domain consistency loss for the generator is introduced in Cross-domain Correspondence (CDC) \cite{ojha2021few-shot-gan} to preserve pairwise distance information learned from source domains during adaptation. Moreover, CDC adds patch-level discrimination to relieve overfitting. The patch-level discrimination judges realism based on intermediate features of the discriminator, where the receptive field of each member corresponds to a patch in the input image. Nevertheless, CDC still fails to preserve the diversity of some detailed characteristics, like detailed hairstyles and facial expressions of humans. 

    In this work, we focus on regulating the discriminator, which easily overfits and guides the generator to replicate training samples in few-shot image generation tasks. As shown in Fig. \ref{begin2}, our approach guides adapted GANs to learn from target domains with limited data and allows them to preserve diverse information learned from source domains, improving generation quality and diversity. We propose masked discrimination to further relieve overfitting and achieve more diverse results. We apply random masks to features extracted by the target discriminator, forcing it to judge realism with partially masked features. As a result, the target generator is guided to generate diverse images sharing partially common features with real samples, leading to greater diversity. Besides, we employ the cross-domain consistency loss for the discriminator to achieve higher generation quality by strengthening global image discrimination and preserving information learned from source domains. The discriminator cross-domain consistency loss encourages the target discriminator to extract diverse features from input images and keep relative distances between generated samples in its feature space similar to the source discriminator.

    The main contributions of the proposed few-shot image generation approach are as follows:  (1) We propose masked discrimination to guide adapted GANs to generate images sharing partially common features with training samples and achieve greater diversity. (2) We propose to adapt the cross-domain consistency loss to the discriminator to preserve information learned from source domains and achieve higher quality. (3) We demonstrate the effectiveness of our approach on a series of few-shot image generation tasks with qualitative and quantitative results outperforming prior works.

\section{Related Work}

\textbf{Few-shot image generation}
Few-shot image generation aims to generate diverse and high-quality images using only a few available samples. Most existing works follow the adaptation method proposed in TGAN \cite{wang2018transferring} to adapt GANs pre-trained on large source domains, including ImageNet \cite{deng2009imagenet}, FFHQ \cite{Karras_2019_CVPR}, and LSUN \cite{yu2015lsun} et al., to target domains with limited data. Augmentation approaches \cite{tran2021data,zhao2020differentiable,zhao2020image} like ADA \cite{ada} can help provide more different augmented samples to relieve overfitting. BSA \cite{noguchi2019image} updates the scale and shift parameters in the generator only during adaptation. FreezeD \cite{mo2020freeze} freezes the lower layers of the discriminator to relieve overfitting. EWC \cite{ewc} applies elastic weight consolidation to regularize the generator by making it harder to change critical weights which have higher Fisher information \cite{2017A} values during adaptation. MineGAN \cite{wang2020minegan} adds additional fully connected networks to modify noise inputs for the generator, aiming to shift the distributions of latent space for adaptation. CDC \cite{ojha2021few-shot-gan} builds a correspondence between source and target domains with the cross-domain consistency loss for generators and patch-level discrimination. DCL \cite{zhao2022closer} uses contrastive learning to maximize the similarity between corresponding image pairs in the source and target domain and push away the generated samples from real images for greater diversity. Our approach is compared with the abovementioned approaches in quality and diversity to prove its effectiveness on few-shot training data.

In addition, recent works in few-shot image generation have provided different research perspectives. RSSA \cite{xiao2022few} proposes a relaxed spatial structural alignment method with compressed latent space based on inverted GANs \cite{Abdal_2020_CVPR}. RSSA aims to preserve image structures learned from source domains, which is inappropriate for some abstract target domains like artists' paintings. AdAM \cite{adaptative} introduces an adaptation-aware approach that works especially well for unrelated source/target domains. Researches including MTG \cite{zhu2021mind}, OSCLIP \cite{kwon2022one}, GDA\cite{zhang2022generalized}, and DIFA \cite{zhang2022towards}, et al. focus on exploring single-shot GAN adaptation with the help of pre-trained CLIP \cite{pmlr-v139-radford21a} image encoders.

\textbf{Domain translation}
Domain translation research has provided many approaches based on conditional GANs \cite{isola2017image,zhu2017unpaired,zhu2017toward} and variational autoencoders (VAEs) \cite{lee2018diverse} to convert an image from source domains to target domains. However, most domain translation approaches  \cite{ma2018exemplar,richardson2021encoding} ask for abundant training data in both source and target domains, constraining its application to few-shot tasks. Recent works \cite{pang2021image,saito2020coco} separate the content and style in image translation to relieve this problem. SEMIT \cite{2020wang} applies semi-supervised learning with a noise-tolerant pseudo-labeling procedure. Despite that, abundant class or style-labeled data are still required for conditional GANs or VAEs training. This paper aims to realize model-level unconditional few-shot image generation based on models pre-trained on large-scale source domains and only a few real samples instead of image-level translation.

\section{Approach}
Few-shot image generation aims to achieve high quality and great diversity generation with a few real samples utilizing the source GAN model, which consists of the source generator $G_s$ and source discriminator $D_s$. However, adapted GANs directly fine-tuned on limited data overfit seriously and tend to replicate real samples since the target discriminator $D_t$ can remember all of them. 

The proposed approach focuses on adjusting the optimization target of adapted GANs by regulating the target discriminator $D_{t}$. We follow prior works \cite{ojha2021few-shot-gan} to add patch-level discrimination, which calculates adversarial loss using certain intermediate features of the target discriminator $D_{t}$. The full target discriminator $D_{t}$ is used for input noises sampled from a subset of the latent space $Z_{sub}$ and the patch-level discrimination is applied to the whole latent space. More details of the patch-level discrimination are added in Appendix \ref{appendixa}.

The main contribution of our approach is to propose masked discrimination, which further relieves the overfitting of discriminators. More specifically, we propose to apply random masks to features extracted by the target discriminator $D_{t}$. The target discriminator $D_{t}$ is encouraged to judge more diverse images sharing partially common features with training samples as realistic for greater generation diversity (Sec \ref{sec2}). To further preserve information learned from source domains, we adapt the cross-domain consistency loss (CDC loss) \cite{ojha2021few-shot-gan}, which is originally designed for generators, to regularize the target discriminator $D_t$ for higher generation quality (Sec \ref{sec1}). The target generator $G_t$ is trained to learn from the limited real samples and preserve the diversity provided by pre-trained source GANs under the guidance of the target discriminator $D_t$. 

\subsection{Masked discrimination}
\label{sec2}
\begin{figure}[t]
    \centering
    \includegraphics[width=1.0\linewidth]{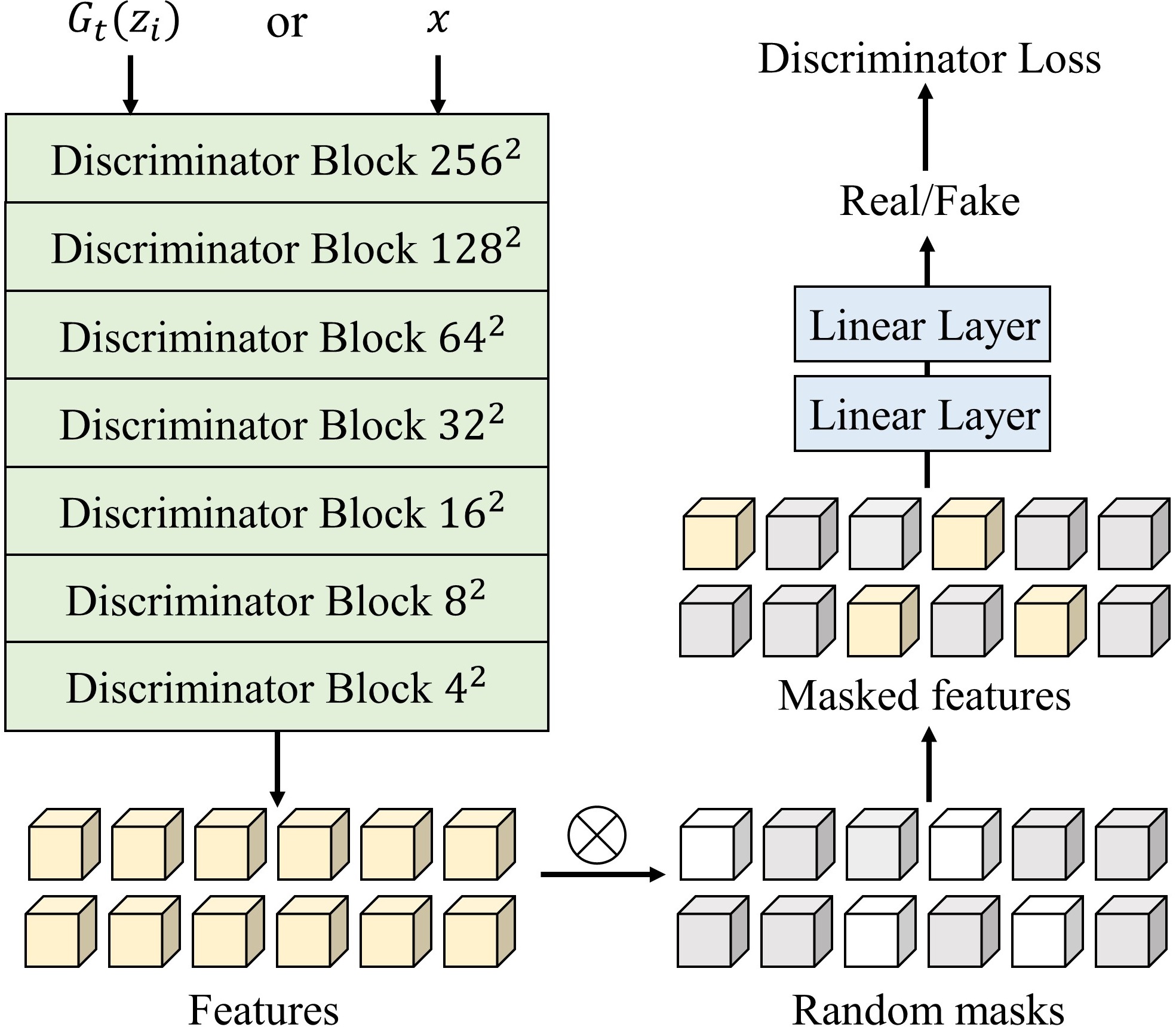}
    \caption{We apply random masks to features extracted by the target discriminator $D_{t}$ from input images for greater generation diversity using image resolution $256\times 256$ and 2/3 features masked (marked as gray) as an example. }
    \label{masked}
\end{figure}

To prevent the target discriminator $D_{t}$ from remembering all the training samples and guiding the target generator $G_{t}$ to replicate them in few-shot image generation tasks, we apply random masks to features extracted by the target discriminator $D_{t}$ from the input fake generated images $G_t(z_i)$ or real images $x$ as shown in Fig. \ref{masked}. The target discriminator $D_{t}$ is trained to judge the realism of input images based on the randomly masked features. In this way, it becomes difficult for $D_{t}$ to remember any training samples. As a result, $D_t$ is encouraged to judge generated samples sharing partially common features with training samples as realistic instead of pursuing replications of training samples. Correspondingly, the target generator $G_t$ can produce more diverse images different from the training samples, slowing down diversity degradation during adaptation.

We apply random masks to the output features of different discriminator blocks and the first following linear layer and achieve the most realistic results with great diversity when masking the output features of discriminator block $4^2$. Besides, we use different sizes of masks ranging from $1/2$ to $3/4$ of the feature space size and achieve diverse results as illustrated in Sec \ref{results}. Detailed ablations of the masked layer choice and mask size are provided in Appendix \ref{ablationslambdamask}.

While masked discrimination improves the generation diversity of adapted GANs, it weakens discrimination on the global image and may lead to degraded quality. Therefore, we propose discriminator CDC loss to better preserve the information learned from source domains for higher generation quality.

\subsection{Discriminator CDC loss}
\label{sec1}

\begin{figure}[tbp]
    \centering
    \includegraphics[width=1.0\linewidth]{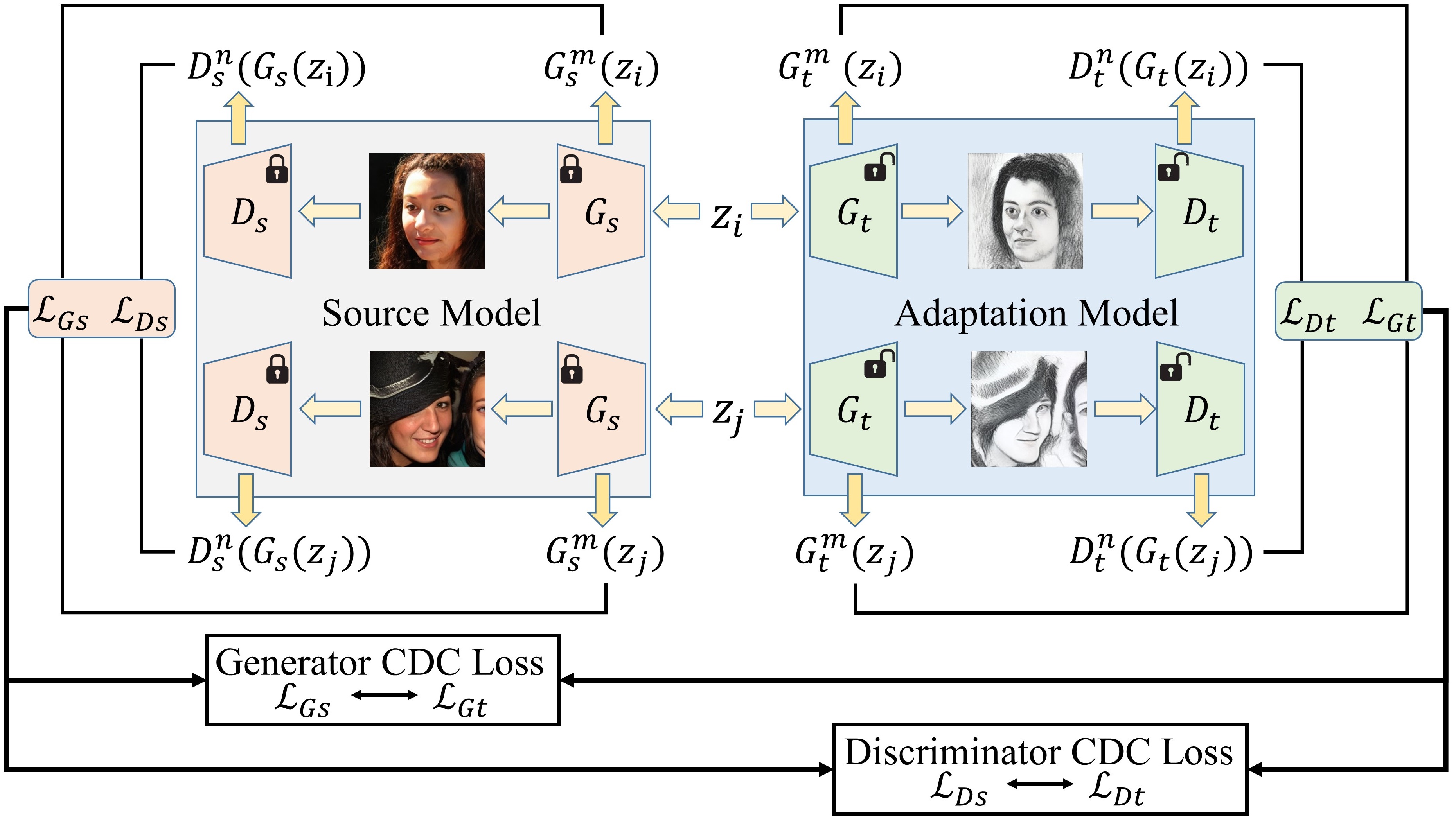}
    \caption{We apply the cross-domain consistency loss to both the generator and discriminator for higher generation quality. $\mathcal{L}_{Gs}$ and $\mathcal{L}_{Ds}$ represent the relative distances between two samples in the feature space of the source generator and discriminator. Similarly, $\mathcal{L}_{Gt}$ and $\mathcal{L}_{Dt}$ represent those of the target generator and discriminator.}
    \label{cdc}
\end{figure}

As shown in Fig. \ref{cdc}, we propose to apply the cross-domain consistency loss to both the generator and discriminator, aiming to keep pairwise relative distances between generated samples and preserve information provided by source GANs for higher image quality. Given two arbitrary input noises $z_i,z_j$ for the generator, the pairwise relative distances in the feature space of the source and target generator can be given by $sim(G_s^m(z_i),G_s^m(z_j))$ and $sim(G_t^m(z_i),G_t^m(z_j))$, where $sim$ represents the cosine similarity between activations at the $m^{th}$ layer of the generators. Similarly, we have pairwise relative distances in the feature space of the source and target discriminator as $sim(D_s^n(G_s(z_i)),D_s^n(G_s(z_j)))$ and $sim(D_t^n(G_t(z_i)),D_t^n(G_t(z_j)))$, where $sim$ represents the cosine similarity between activations at the $n^{th}$ layer of the discriminators. A batch of $K+1$ noise vectors $\left\lbrace z_k \right\rbrace_{0}^{K}, (0\leq k \leq K)$ is needed to construct $K$-way probability distributions for an arbitrary input noise $z_i$ in the source and target generator as follows:

\begin{equation}
\begin{aligned}
     p_{g,i}^{s,m} &= Softmax(\left\lbrace sim(G_s^m(z_i),G_s^m(z_j)) \right\rbrace_{\forall i\neq j}), \\
    p_{g,i}^{t,m} &= Softmax(\left\lbrace sim(G_t^m(z_i),G_t^m(z_j)) \right\rbrace_{\forall i\neq j}).
\end{aligned}
\end{equation}

Similarly, the probability distributions in the source and target discriminator are as follows:

\begin{equation}
\begin{aligned}
     p_{d,i}^{s,n} &= Softmax(\left\lbrace sim(D_s^n(G_s(z_i)),D_s^n(G_s(z_j))) \right\rbrace_{\forall i\neq j}), \\
    p_{d,i}^{t,n} &= Softmax(\left\lbrace sim(D_t^n(G_t(z_i)),D_t^n(G_t(z_j))) \right\rbrace_{\forall i\neq j}).
\end{aligned}
\end{equation}

Finally, we have the cross-domain consistency loss $L_{dist}$ across layers and image instances using KL-divergence ($D_{KL}$) for the generator and discriminator as:

\begin{equation}
    \begin{aligned}
         \mathcal{L}_{dist}(G_s,G_t) &= \mathbb{E}_{\left\lbrace z_i\sim p_z(z)\right\rbrace}\sum_{m,i}D_{KL}(p_{g,i}^{t,m}||p_{g,i}^{s,m}), \\
         \mathcal{L}_{dist}(D_s,D_t) &= \mathbb{E}_{\left\lbrace z_i\sim p_z(z)\right\rbrace}\sum_{n,i}D_{KL}(p_{d,i}^{t,n}||p_{d,i}^{s,n}),
    \end{aligned}
\end{equation}
where $p_z(z)$ represents the distributions of input noises. We apply discriminator CDC loss to encourage the target discriminator to extract diverse features like the source discriminator instead of only focusing on features of a few real samples. In this way, the generator is guided to preserve more diverse characteristics learned from source domains during adaptation. Combined with masked discrimination, our full approach achieves more diversified generation results and preserves the information in source GANs better.

\begin{figure*}[t]
    \centering
    \includegraphics[width=1.0\linewidth]{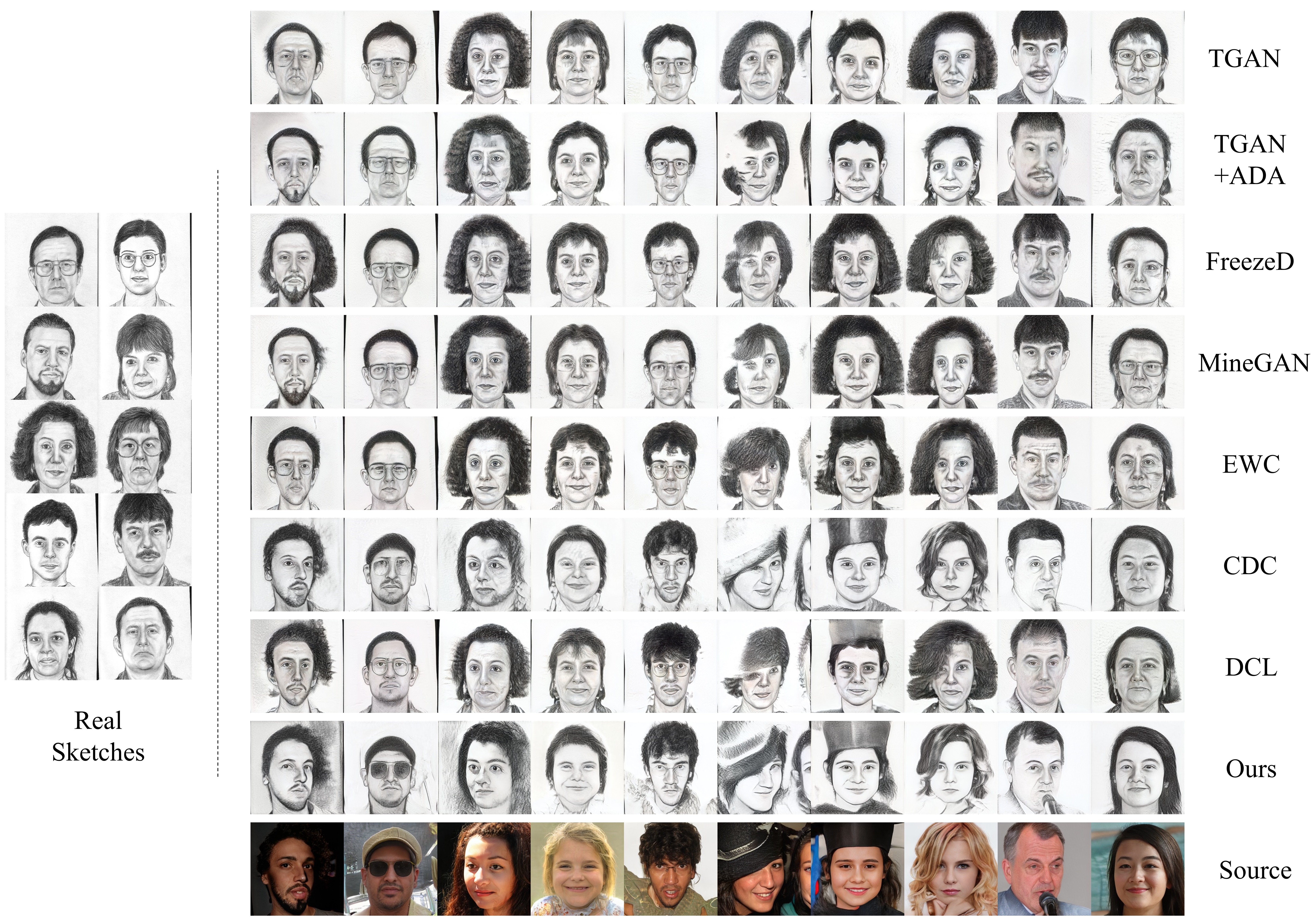}
    \caption{10-shot image generation results on FFHQ $\rightarrow$ Sketches of our approach and baselines. All the visualized samples of different approaches are synthesized from fixed noise inputs for comparison. Our approach achieves higher-quality results and preserves more diverse information naturally from the source domain.}
    \label{sketches}
\end{figure*}

\subsection{Overall optimization target}
Given $x$ and $z$ representing real samples following the distribution $p_{data}(x)$ and random input noises respectively, the non-saturating loss for GANs consisting of the generator $G$ and discriminator $D$ can be expressed as:
\begin{equation}
    \mathcal{L}_{adv}(G,D) = D(G(z)) -D(x).
\end{equation}

The overall loss function of the proposed approach consists of the adversarial loss using image-level and patch-level discrimination and cross-domain consistency loss for the generator and discriminator:
\begin{equation}
\begin{aligned}
     \mathcal{L} &= \mathbb{E}_{x\sim p_{data}(x)}[\mathbb{E}_{z\sim Z_{sub}} \mathcal{L}_{adv}(G_t,D_{t}) \\ 
     &+ \mathbb{E}_{z\sim p_z(z)} \mathcal{L}_{adv}(G_t,D_{p})] \\ &+ \lambda (\mathcal{L}_{dist}(G_s,G_t)+\mathcal{L}_{dist}(D_s,D_t)).
\end{aligned}
\label{loss}
\end{equation}
 Here $D_{p}$ represents a subset of the target discriminator $D_{t}$ used for patch-level discrimination. $D_{p}$ uses intermediate features of $D_{t}$, where the receptive field of each member corresponds to a patch in the input image. The proposed masked discrimination has influences on both image-level and patch-level discrimination. We find $\lambda$ between $10^3$ to $5\times 10^3$ appropriate for adaptation setups in this paper empirically. Detailed ablations of $\lambda$ are added in Appendix \ref{ablationslambdamask}.

\section{Experiments}

\textbf{Basic setups}
We follow the experimental setups used in prior works \cite{mo2020freeze,wang2020minegan,ewc,ojha2021few-shot-gan,zhao2022closer} to implement the proposed approach based on StyleGAN2 \cite{Karras_2020_CVPR}. We adapt pre-trained source GANs to target domains with batch size 4 on a single NVIDIA TITAN RTX GPU. Our approach is mainly compared with prior works on 10-shot adaptation tasks qualitatively and quantitatively. Additional results on 5-shot and 1-shot adaptation tasks are added in Appendix \ref{appendixe}. 

\begin{figure*}[t]
    \centering
    \includegraphics[width=1.0\linewidth]{ 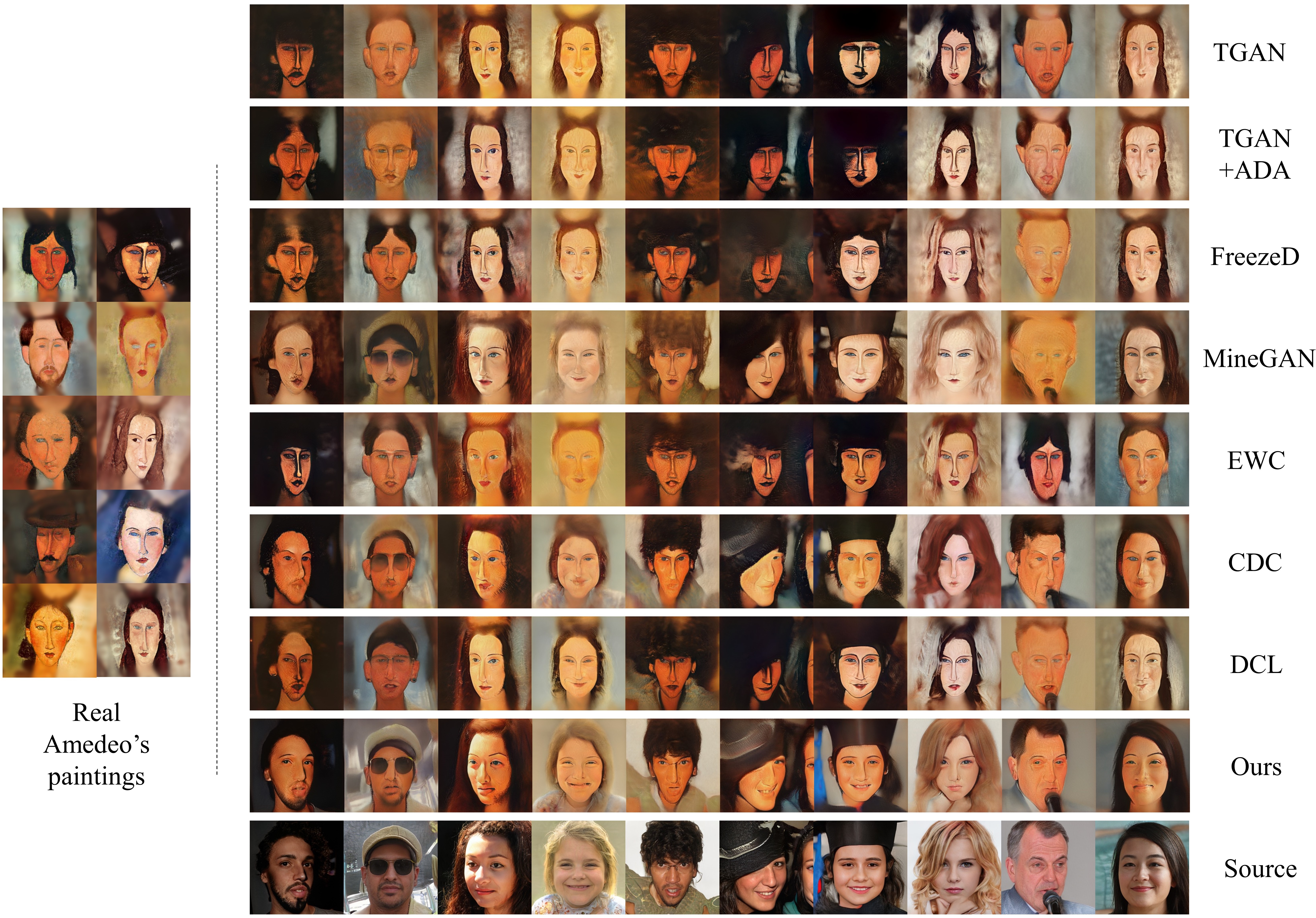}
    \caption{10-shot image generation results on FFHQ $\rightarrow$ Amedeo's paintings of our approach and baselines. All the visualized samples of different approaches are synthesized from fixed noise inputs for comparison. Our approach is capable of learning the common features of a few real samples and preserves diverse characteristics like sunglasses, hats, hairstyles, and facial expressions.}
    \label{amedeo}
\end{figure*}

\textbf{Baselines}
We employ 7 baselines sharing similar targets with us to adapt source GANs to target domains with only a few available samples for comparison: TGAN \cite{wang2018transferring}, TGAN+ADA \cite{ada}, FreezeD \cite{mo2020freeze}, MineGAN \cite{wang2020minegan}, EWC \cite{ewc}, CDC \cite{ojha2021few-shot-gan}, and DCL \cite{zhao2022closer}. We do not use BSA \cite{noguchi2019image} proposed based on BigGAN \cite{DBLP:conf/iclr/BrockDS19} for comparison since, in most cases, it cannot achieve better results for StyleGAN2-based models than TGAN \cite{wang2018transferring}, which directly fine-tunes source GANs on target domains using limited data.

\textbf{Datasets}
StyleGAN2 \cite{Karras_2020_CVPR} models pre-trained on large-scale datasets including FFHQ \cite{Karras_2019_CVPR}, LSUN Church, and LSUN Cars \cite{yu2015lsun} are used as source models. We evaluate the proposed approach with few-shot adaptation to various target domains, including Sketches \cite{wang2008face}, FFHQ-Babies (Babies), FFHQ-Sunglasses (Sunglasses) \cite{Karras_2019_CVPR}, face paintings by Amedeo Modigliani, Raphael Peale, and Otto Dix \cite{yaniv2019face}, Haunted houses, Van Gogh houses, and Wrecked cars. The training images are resized to the resolution of $256\times 256$ except for those from LSUN Cars and Wrecked cars, which are resized to $512\times 512$. Real samples of all the target datasets can be found in Appendix \ref{appendixe}.

\begin{figure*}[t]
    \centering
    \includegraphics[width=1.0\linewidth]{ 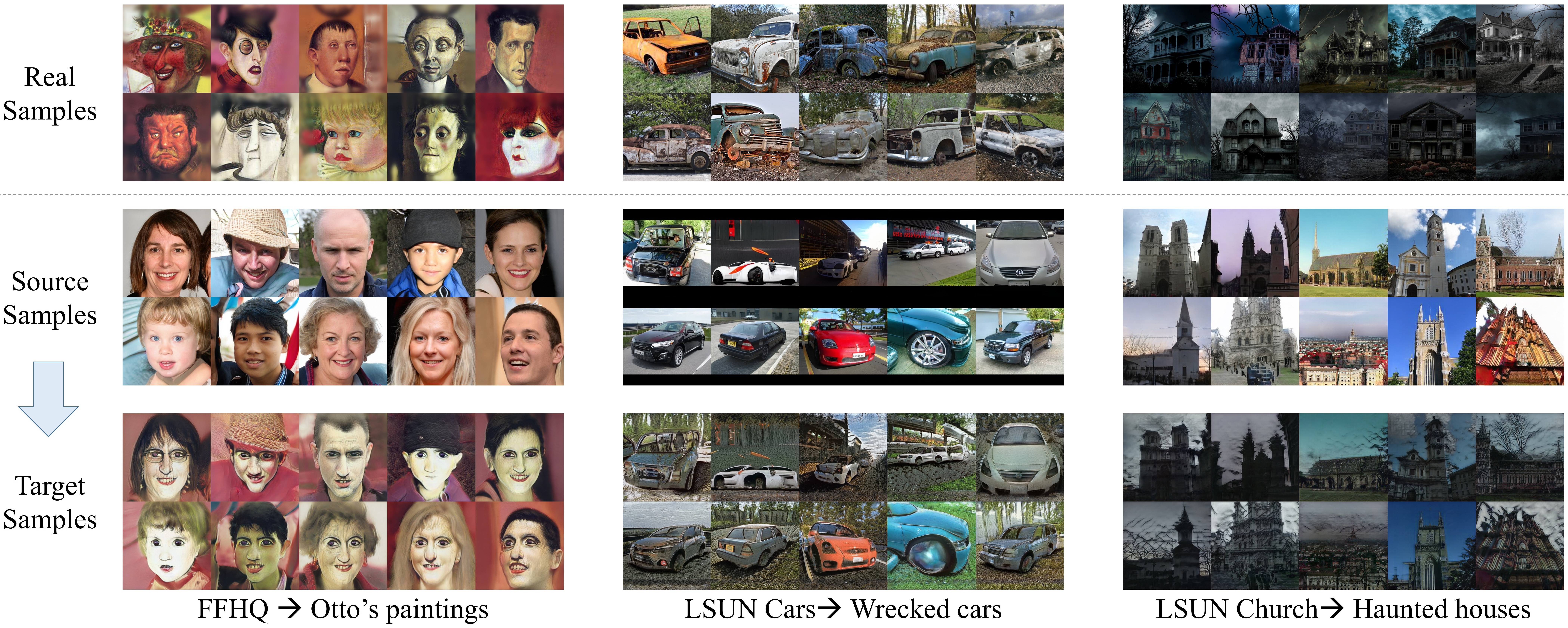}
    \caption{Additional 10-shot image generation results of our approach under different adaptation setups including FFHQ $\rightarrow$ Otto's paintings, LSUN Cars $\rightarrow$ Wrecked cars, and LSUN Church $\rightarrow$ Haunted houses.}
    \label{result2}
\end{figure*}

\begin{table*}[tbp]
\setlength\tabcolsep{4.8pt}
\centering
\small
\begin{tabular}{l|c|c|c|c|c|c}
   Approaches & \makecell[c]{FFHQ $\rightarrow$ \\ Sketches} & \makecell[c]{FFHQ $\rightarrow$ \\ Babies} & \makecell[c]{FFHQ $\rightarrow$ \\ Sunglasses} & \makecell[c]{FFHQ $\rightarrow$ \\ Otto's paintings} & \makecell[c]{FFHQ $\rightarrow$ \\ Raphael’s paintings} & \makecell[c]{FFHQ $\rightarrow$ \\ Amedeo’s paintings}  
 \\
\hline
TGAN \cite{wang2018transferring} & $0.394 \pm 0.023$ & $0.510 \pm 0.026$ & $0.550 \pm 0.021$ & $0.594 \pm 0.023$ & $0.533 \pm 0.023$ & $0.548 \pm 0.026$  \\
TGAN+ADA \cite{ada} & $0.427 \pm 0.022$ & $0.546 \pm 0.033$ & $0.571 \pm 0.034$ & $0.625 \pm 0.028$ & $0.546 \pm 0.037$ & $0.560 \pm 0.019$ \\
FreezeD \cite{mo2020freeze} & $0.406 \pm 0.017$ & $0.535 \pm 0.021$ & $0.558 \pm 0.024$ & $0.629 \pm 0.023$ & $0.537 \pm 0.026$ & $0.558 \pm 0.019$ \\
MineGAN \cite{wang2020minegan} & $0.407 \pm 0.020$ & $0.514 \pm 0.034$ & $0.570 \pm 0.020$ & $0.625 \pm 0.030$ & $0.559 \pm 0.031$ & $0.586 \pm 0.041$ \\
EWC \cite{ewc} & $0.430 \pm 0.018$ & $0.560 \pm 0.019$ & $0.550 \pm 0.014$ & $0.611 \pm 0.025$ & $0.541 \pm 0.023$ & $0.579 \pm 0.035$ \\
CDC \cite{ojha2021few-shot-gan} & $0.454 \pm 0.017$ & $0.583 \pm 0.014$ & $0.581 \pm 0.011$ & $\pmb{0.638 \pm 0.023}$ & $0.564 \pm 0.010$ & $0.620 \pm 0.029$ \\
DCL \cite{zhao2022closer} & $0.461 \pm 0.021$ & $0.579 \pm 0.018$ & $0.574 \pm 0.007$ & $0.617 \pm 0.033$ & $0.558 \pm 0.033$ & $0.616 \pm 0.043$ \\
Ours & $\pmb{0.505 \pm 0.020}$ & $\pmb{0.595 \pm 0.006}$ & $\pmb{0.593 \pm 0.014}$ & $\pmb{0.638 \pm 0.024}$ & $\pmb{0.581 \pm 0.012}$ & $\pmb{0.628 \pm 0.024}$ \\
\end{tabular}
\caption{Intra-LPIPS ($\uparrow$) results of 10-shot image generation tasks adapted from the source domain FFHQ. Standard deviations are computed across 10 clusters (the same number as training samples). }
\label{lpipsffhq}
\end{table*}

\textbf{Evaluation metrics}
The main purpose of the proposed approach is to generate high-quality images with great diversity based on only a few available samples from target domains. Therefore, we employ FID \cite{heusel2017gans} and Intra-LPIPS \cite{ojha2021few-shot-gan} to evaluate the quality and diversity of generation results, respectively. 

FID is widely applied to evaluate the ability of generators to reproduce real distributions using the distribution distance between generated results and real data. However, FID becomes unstable and unreliable for datasets containing only a few samples. Therefore, we use datasets containing relatively abundant data for stable FID evaluation.

Intra-LPIPS is based on LPIPS \cite{zhang2018unreasonable} which evaluates the perceptual distances \cite{johnson2016perceptual} between images. We follow settings in prior works \cite{ojha2021few-shot-gan} to calculate Intra-LPIPS. We generate 1000 images with the adapted GAN and assign them to one of the training samples with the closest perceptual distance (lowest LPIPS value). The Intra-LPIPS metric can be calculated with the average pairwise LPIPS within every cluster averaged over all the clusters. With exactly replicated training samples, the Intra-LPIPS metric will have a score of zero. More diverse generated results correspond to larger Intra-LPIPS values. We calculate Intra-LPIPS with fixed input noises to fairly compare our approach's generation diversity with baselines on 10-shot adaptation tasks.

\subsection{Quality and diversity evaluation}
\label{results}
In this section, we provide a comprehensive evaluation of our approach both quantitatively and qualitatively. The proposed approach is compared with baselines through visualized samples and evaluation metrics, including FID and Intra-LPIPS, under different adaptation setups.

\begin{figure*}[tbp]
    \centering
    \includegraphics[width=1.0\linewidth]{ 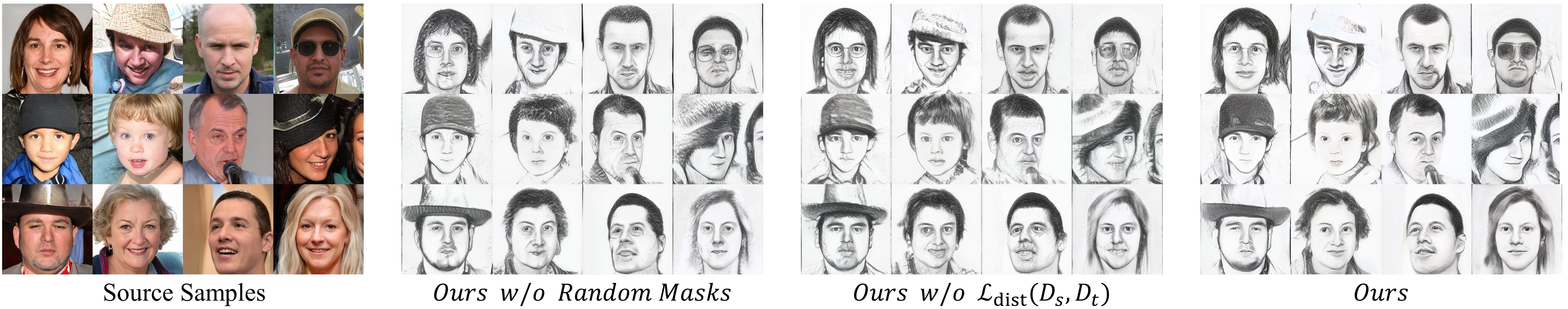}
    \caption{Ablations of the proposed approach on 10-shot FFHQ $\rightarrow$ Sketches. Absence of random masks leads to unnatural adaptation of characteristics like hairstyles, occlusion of hats and sunglasses, and facial expressions. Absence of discriminator CDC loss makes adapted GANs produce images containing blurs, leading to degraded image quality.}
    \label{ablations}
\end{figure*}

\textbf{Qualitative evaluation}
We visualize the few-shot image generation results on 10-shot FFHQ $\rightarrow$ Sketches and FFHQ $\rightarrow$ Amedeo's paintings of our approach and baselines in Fig. \ref{sketches} and \ref{amedeo} and provide supplemental results under other adaptation setups in Appendix \ref{appendixe}. For fair comparison, we exhibit results generated from fixed input noises for different methods. We find that TGAN and TGAN+ADA overfit seriously and tend to replicate the training samples. FreezeD and EWC add limitations to preserve parameters in the discriminator and generator, respectively. These approaches can produce more diverse results but still replicate some of the training samples and cannot naturally transfer characteristics from source domains to target domains. MineGAN adds additional networks to adjust the distributions of input noises. Like FreezeD and EWC, MineGAN cannot avoid producing some replications of the training samples and gets unnatural results. CDC, DCL, and our approach build a correspondence between the source and target domains for diversity preservation. As shown in the visualized results, our approach generates diverse images and preserves detailed characteristics of source domains during adaptation to target domains. Our approach achieves higher quality and greater diversity than baselines. For example, our approach inherits the occlusion of hats and sunglasses from the source domain more naturally. Moreover, the hairstyles and facial expressions are better preserved as well. More results under different adaptation setups, including FFHQ $\rightarrow$ Otto's paintings, LSUN Cars $\rightarrow$ Wrecked cars, and LSUN Church $\rightarrow$ Haunted houses, are shown in Fig \ref{result2}. Our approach also achieves pleasing visual effects while maintaining diversity under these adaptation setups. Additional visualized results can be found in Appendix \ref{appendixe}.

\begin{table}[tbp]
\centering
\setlength\tabcolsep{3.5pt}
\small
\begin{tabular}{l|c|c|c}
Approaches & Sketches & Babies & Sunglasses \\
\hline
TGAN \cite{wang2018transferring} & $53.42 \pm 0.02$ & $104.79 \pm 0.03$ & $55.61 \pm 0.04$ \\
TGAN+ADA \cite{ada} & $66.99 \pm 0.01$ & $102.58 \pm 0.12$ & $53.64 \pm 0.08$ \\
%BSA & 69.32 \pm 0.02 & 140.34 \pm 0.01 & 76.12 \pm 0.01 \\
FreezeD \cite{mo2020freeze} & $46.54 \pm 0.01$ & $110.92 \pm 0.02$ & $51.29 \pm 0.05$ \\
MineGAN \cite{wang2020minegan} & $64.34 \pm 0.02$ & $98.23 \pm 0.03$ & $68.91 \pm 0.03$ \\
EWC \cite{ewc} & $71.25 \pm 0.01$ & $87.41 \pm 0.02$ & $59.73 \pm 0.04$ \\
CDC \cite{ojha2021few-shot-gan} & $45.67 \pm 0.02$ & $74.39 \pm 0.03$ & $42.13 \pm 0.04$ \\
DCL \cite{zhao2022closer} & $37.90 \pm 0.02$ & $52.56 \pm 0.02$ & $38.01 \pm 0.01$ \\
Ours & $\pmb{28.93 \pm 0.01}$ & $\pmb{36.39 \pm 0.01}$ & $\pmb{26.96 \pm 0.01}$ \\
\end{tabular}
\caption{FID ($\downarrow$) evaluation using target domains containing relatively abundant data (source domain: FFHQ). Standard deviations are computed across 5 runs.}
\label{fid}
\end{table}

\begin{table}[tbp]
\centering
\begin{tabular}{l|c|c}
 Approaches & \makecell[c]{LSUN Church $\rightarrow$ \\ Haunted houses} & \makecell[c]{LSUN Cars $\rightarrow$ \\ Wrecked cars} \\
\hline
TGAN \cite{wang2018transferring} & $0.585 \pm 0.007$ & $0.592 \pm 0.031$ \\
TGAN+ADA \cite{ada} &  $0.615 \pm 0.018$ & $0.618 \pm 0.034$ \\
FreezeD \cite{mo2020freeze} & $0.581 \pm 0.013$ & $0.572 \pm 0.037$ \\
MineGAN \cite{wang2020minegan} &  $0.617 \pm 0.007$ & $0.620 \pm 0.043$ \\
EWC \cite{ewc} & $0.590 \pm 0.014$ & $0.595 \pm 0.021$  \\
CDC \cite{ojha2021few-shot-gan} & $0.623 \pm 0.026$ & $0.614 \pm 0.027$ \\
DCL \cite{zhao2022closer} & $ 0.614 \pm 0.013$ & $0.609 \pm 0.032$ \\
Ours & $\pmb{0.632 \pm 0.020}$ & $\pmb{0.626 \pm 0.034}$\\
\end{tabular}
\caption{Additional Intra-LPIPS ($\uparrow$) results of 10-shot image generation tasks: LSUN Church $\rightarrow$ Haunted houses and LSUN Cars $\rightarrow$ Wrecked cars. Standard deviations are computed across 10 clusters (the same number as training samples).}
\label{lsun}
\end{table}

\textbf{Quantitative evaluation}
To evaluate the capability of our approach to model real distributions, we use datasets containing relatively abundant images for FID evaluation, including the original Sketches, FFHQ-Babies, and FFHQ-Sunglasses datasets which roughly contain 300, 2500, and 2700 images, respectively. Our approach outperforms all the baselines remarkably on all three datasets as shown in Table \ref{fid}, demonstrating its strong ability to reproduce target distributions using limited data.

We evaluate the generation diversity of our approach and other baselines under several 10-shot adaptation setups: FFHQ $\rightarrow$ Sketches; Babies; Sunglasses; Otto's paintings; Raphael's paintings; Amedeo's paintings. We report the Intra-LPIPS results of these adaptation setups in Table \ref{lpipsffhq}. Our approach outperforms baselines on almost all the benchmarks in terms of Intra-LPIPS when adapted from the source domain FFHQ. To further prove the robustness of our approach, we compare it with baselines on the other two 10-shot adaptation tasks: LSUN Church $\rightarrow$ Haunted houses and LSUN Cars $\rightarrow$ Wrecked cars as shown in Table \ref{lsun} and find improved results as well. With masked discrimination, our approach guides the few-shot adapted GANs to generate more diverse images different from training samples. All the Intra-LPIPS results of our approach and baselines are calculated with fixed input noises for fair comparison.

\subsection{Ablations}
\label{ablations_experiments}
Compared with existing few-shot image generation methods, our approach proposes two new ideas, random masks applied to the discriminator and discriminator CDC loss. As shown in Fig \ref{ablations}, we provide visualized ablations of our approach using 10-shot FFHQ $\rightarrow$ Sketches as an example. Masked discrimination encourages the discriminator to judge more diverse images as realistic and guides the adapted GAN to learn the common features of limited training samples. Absence of random masks makes the adapted GAN hard to preserve characteristics like hairstyles and the occlusion of hats and sunglasses, which are obviously different from the training samples. Discriminator CDC loss helps strengthen global image discrimination and preserve information learned from the source domain. Without it, the adapted GAN produces results containing unnatural blurs and artifacts, leading to degraded image quality.  

Moreover, we provide a quantitative ablation analysis in terms of Intra-LPIPS and FID (see Table \ref{qablation} and \ref{qablation2} in Appendix \ref{ablationslambdamask}). Both ideas help improve generation diversity and learning of target distributions, while masked discrimination contributes more to the overall improvement of Intra-LPIPS and FID. Our full approach achieves high-quality few-shot image generation results with great diversity, benefiting from the combination of these two ideas.

\section{Conclusion and Limitations}
This paper proposes to realize few-shot image generation via masked discrimination. Our approach focuses on regulating the discriminator, which easily overfits and guides the generator to replicate training samples in few-shot image generation tasks. We apply random masks to features extracted by the discriminator to encourage the adapted GANs to learn the common features of limited training samples and produce more diverse images different from them. Besides, we propose the discriminator CDC loss to strengthen global image discrimination and achieve higher image quality by preserving information learned from source domains. Our approach is simple and proven effective both qualitatively and quantitatively, with higher generation quality and greater generation diversity than prior works.

While our approach has achieved compelling results, it is not without limitations. For example, our approach may preserve features inappropriate for target domains like beards and wrinkles for babies, as shown in Fig. \ref{babies} (10-shot FFHQ $\rightarrow$ Babies, Appendix \ref{appendixe}). It remains a challenge for adapted GANs to discriminate what kind of features should be preserved or modified with only a few available real samples. Nevertheless, our work has made significant progress in few-shot image generation and can generate high-quality and diverse images. We hope that our work will be a solid basis for better methods in the future.

\bibliographystyle{abbrv}
\bibliography{refs}

\clearpage

\appendix

\section{More Details of Implementation}
\label{appendixa}

\textbf{Our approach} We implement the proposed approach based on StyleGAN2 \cite{Karras_2020_CVPR} following prior works. StyleGAN2 models pre-trained on FFHQ, LSUN Church, and LSUN Cars are used as source models. In addition, we employ LSUN Horses for experiments on unrelated source/target domains (see Fig. \ref{unrelated} in Appendix \ref{appendixe}). The image resolution of all datasets are $256\times 256$ except for LSUN Cars and its corresponding target domain, Wrecked cars, whose resolution are $512\times 512$. The size of features output from discriminator block $4^2$ is 8192. We randomly mask the output features without grouping. We visualize output features as a group of cubes in Fig. \ref{masked} to show the mask ratio clearly. For quantitative evaluation results listed in Table \ref{lpipsffhq}, \ref{fid},  and \ref{lsun}, we randomly mask 3/4 features extracted by the discriminator and set the weight coefficient of cross-domain consistency loss $\lambda$ as 1000. The learning rate is set as 0.002. Adam optimizer \cite{kingma2014adam} is used for updating trainable parameters. With a series of experiments, we find that most of the adaptation setups used in our paper need about 1000 to 1500 iterations to achieve good results (e.g., we use 1500 iterations on FFHQ $\rightarrow$ Sketches for our approach and baselines). More iterations are needed to adapt source GANs to target domains for complex adaptation cases like LSUN Cars $\rightarrow$ Wrecked cars. As shown in Fig. \ref{result2}, samples from LSUN Cars have black upper and lower boundaries, which do not exist in samples of Wrecked cars. We train the source model for 3000 iterations to achieve good results for this adaptation case.

Here, we add more details of the patch-level discrimination \cite{ojha2021few-shot-gan}. The patch-level discrimination makes use of certain intermediate features of the target discriminator for adversarial loss calculation. Our experiments follow prior works to use effective patch sizes ranging from $22 \times 22$ to $61 \times 61$ for patch-level discrimination. Patch-level discrimination is applied to the whole latent space while the full target discriminator (image-level discrimination) is applied to noises sampled from $Z_{sub}$ only. To define the subset of the latent space $Z_{sub}$ for $k$-shot generation, we first select $k$ random points and randomly sample from these fixed points with an added Gaussian noise (standard deviation $\sigma=0.05$) following CDC \cite{ojha2021few-shot-gan}. Noises sampled from $Z_{sub}$ take up 25 percent of all input noises in the training of adapted GANs. 

Since the patch-level discrimination uses the intermediate features of the target discriminator, we do not need to exclude noises from $Z_{sub}$ for the patch-level discrimination. Suppose an image generated from $Z_{sub}$ is judged as realistic by the full target discriminator. In that case, it will be judged as realistic by the patch-level discrimination as well, based on the intermediate features corresponding to the patches in images. 

It is worth noting that the proposed discriminator CDC loss is calculated for both patch-level and image-level discrimination. Therefore, only layers used by both patch-level and image-level discrimination are applied to discriminator CDC loss calculation. The proposed masked discrimination is applied to image-level discrimination rather than patch-level discrimination. Although the full target discriminator only deals with noises sampled from $Z_{sub}$, the adapted GANs still benefit from random masks applied to the target discriminator and achieve greater generation diversity as shown in ablations (Sec \ref{ablations_experiments}) in our paper, since the patch-level discrimination actually serves as a subset of the full target discriminator. The image-level discrimination implemented by the full target discriminator using masked discrimination also influences the patch-level discrimination and guides the generator to achieve greater diversity. 

\textbf{Baselines}
We implement TGAN \cite{wang2018transferring} and TGAN+ADA \cite{ada} based on the implementation of StyleGAN2 \cite{Karras_2020_CVPR}. For FreezeD \cite{mo2020freeze}, we freeze the first 4 layers of StyleGAN2's discriminator following ablations of fixed layers in their work. For MineGAN \cite{wang2020minegan} and CDC \cite{ojha2021few-shot-gan}, results are produced from their official implementations. Since EWC \cite{ewc} and DCL \cite{zhao2022closer} have no official implementation, we implement these approaches following formulas and hyperparameters provided in the papers. All the baselines and our approach are implemented based on the same StyleGAN2 \cite{Karras_2020_CVPR} codebase.

\begin{table}[t]
\centering
\setlength\tabcolsep{4pt}
\begin{tabular}{l|c|c}
 Masked Layer & \makecell[c]{FFHQ $\rightarrow$ \\ Babies} & \makecell[c]{FFHQ $\rightarrow$ \\ Sunglasses} \\
\hline
discriminator block $256^2$ & $0.582 \pm 0.020$ & $0.581 \pm 0.012$ \\
discriminator block $128^2$ & $0.572 \pm 0.014$ & $0.576 \pm 0.016$ \\
discriminator block $64^2$ & $0.575 \pm 0.012$ & $0.572 \pm 0.014$ \\
discriminator block $32^2$ & $0.588 \pm 0.017$ & $0.560 \pm 0.009$  \\
discriminator block $16^2$ & $0.590 \pm 0.022$ & $0.574 \pm 0.019$  \\
discriminator block $8^2$ & $0.590 \pm 0.018$ & $0.576 \pm 0.011$ \\
discriminator block $4^2$ & $\pmb{0.595 \pm 0.006}$ & $\pmb{0.593 \pm 0.014}$ \\
first linear layer & $0.590 \pm 0.013$ &  $0.584 \pm 0.017$ \\
\end{tabular}
\caption{Quantitative ablations of the masked layer on 10-shot FFHQ $\rightarrow$ Babies and FFHQ $\rightarrow$ Sunglasses in terms of Intra-LPIPS ($\uparrow$). Standard deviations are computed across 10 clusters (the same number as training samples).}
\label{ablation_layer}
\end{table}

\section{Supplemental Ablations}
\label{ablationslambdamask}

\begin{figure*}[tbp]
    \centering
    \includegraphics[width=1.0\linewidth]{ 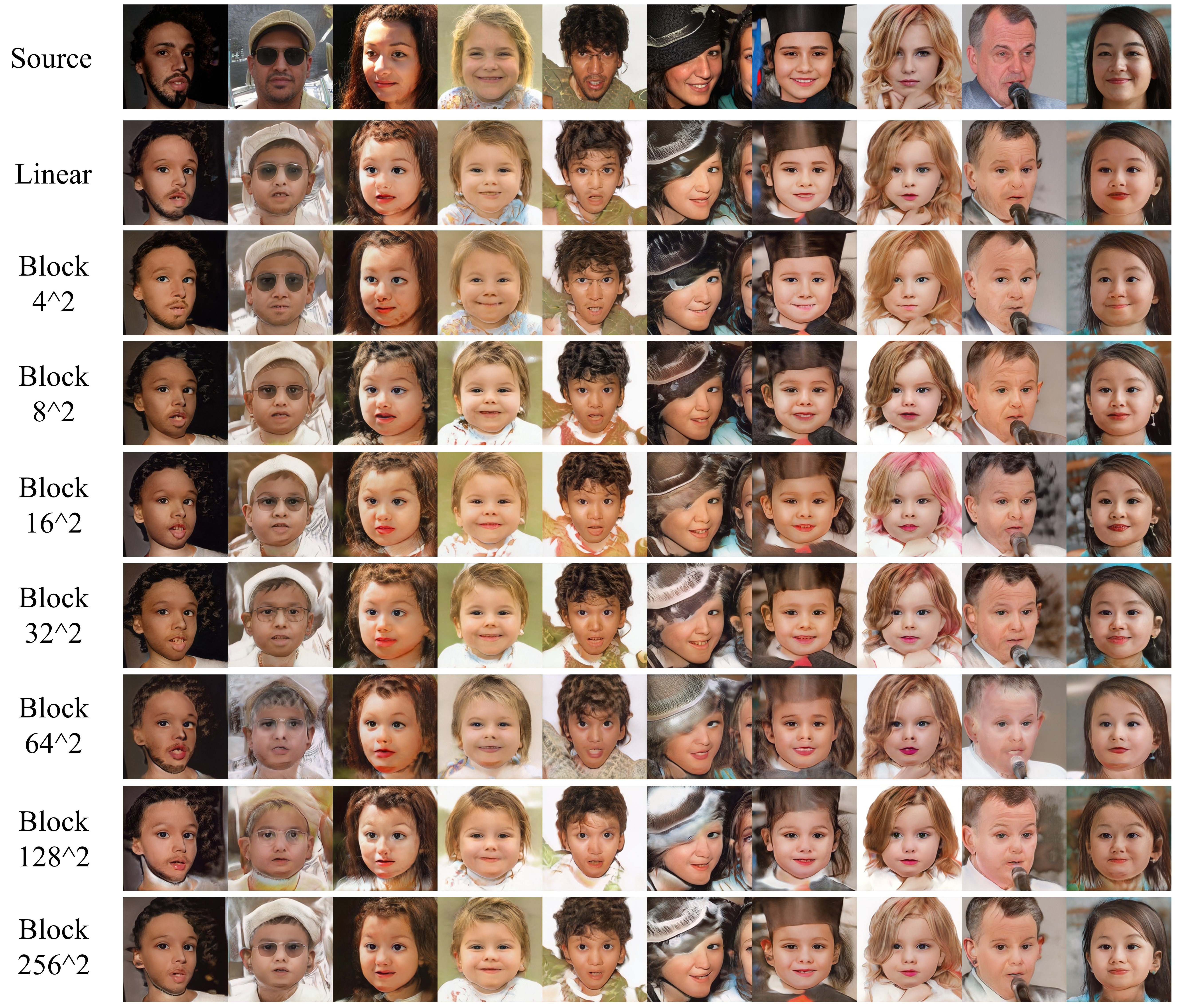}
    \caption{Visualized ablations of the masked layer in the target discriminator on 10-shot FFHQ $\rightarrow$ Babies. }
    \label{layerablation}
\end{figure*}

\begin{figure*}[tbp]
    \centering
    \includegraphics[width=1.0\linewidth]{ 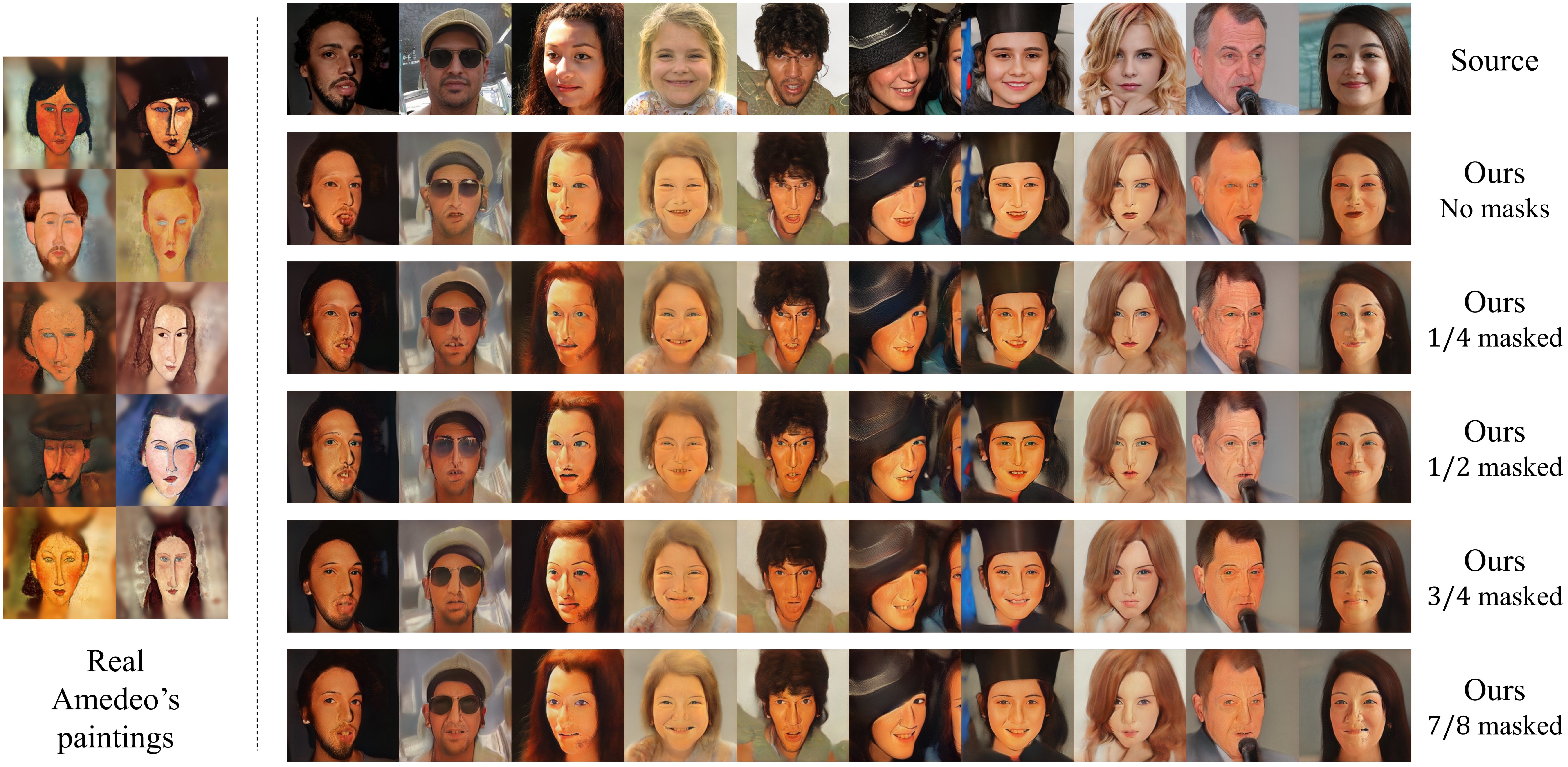}
    \caption{Visualized ablations of the mask size applied to features extracted by the target discriminator (mask ratio of features) on 10-shot FFHQ $\rightarrow$ Amedeo's paintings.}
    \label{maskablation}
\end{figure*}

\textbf{Ablations of masked layer}
We propose masked discrimination to apply random masks to the feature maps output by discriminator block $4^2$ as shown in Fig. \ref{masked}. We tried to apply random masks to the outputs of different layers of the discriminator and provide qualitative and quantitative evaluation as follows. As shown in Table \ref{ablation_layer}, the approach used in the paper consistently, which masks the output of discriminator block $4^2$, achieves the best Intra-LPIPS results. Moreover, we provide visualized comparison in Fig. \ref{layerablation}. Our approach preserves more diverse information (e.g., hairstyles, sunglasses, and facial expressions) and achieves better generation quality. Moreover, masking the output of discriminator block $4^2$ also avoids influencing the calculation of discriminator CDC loss, which makes use of features from discriminator blocks $32^2$ and $16^2$ to preserve more information learned from source domains.

\textbf{Ablations of mask size} In our work, we propose to apply random masks to features extracted by the target discriminator, encouraging adapted GANs to generate more diverse images. As shown in Fig. \ref{maskablation}, we use different sizes of masks on 10-shot FFHQ $\rightarrow$ Amedeo's paintings for ablation analysis. With larger masks, the adapted GANs tend to preserve more information learned from the source domain and produce images more different from the training samples. For example, faces occupy the maximum ratio of the whole image in results generated from the adapted GAN using $7/8$ masked features. In contrast, most of the training samples only occupy a small part. With too small masks, it is hard for adapted GANs to naturally transfer characteristics of source images, such as facial expressions, hairstyles, and sunglasses, to target domains and preserve diversity. We recommend mask ratios ranging from $1/2$ to $3/4$ for adaptation setups in our paper. The weight coefficient of cross-domain consistency loss ($\lambda$) is set as 2500 in the ablations of mask size.

\begin{figure*}[tbp]
    \centering
    \includegraphics[width=0.95\linewidth]{ 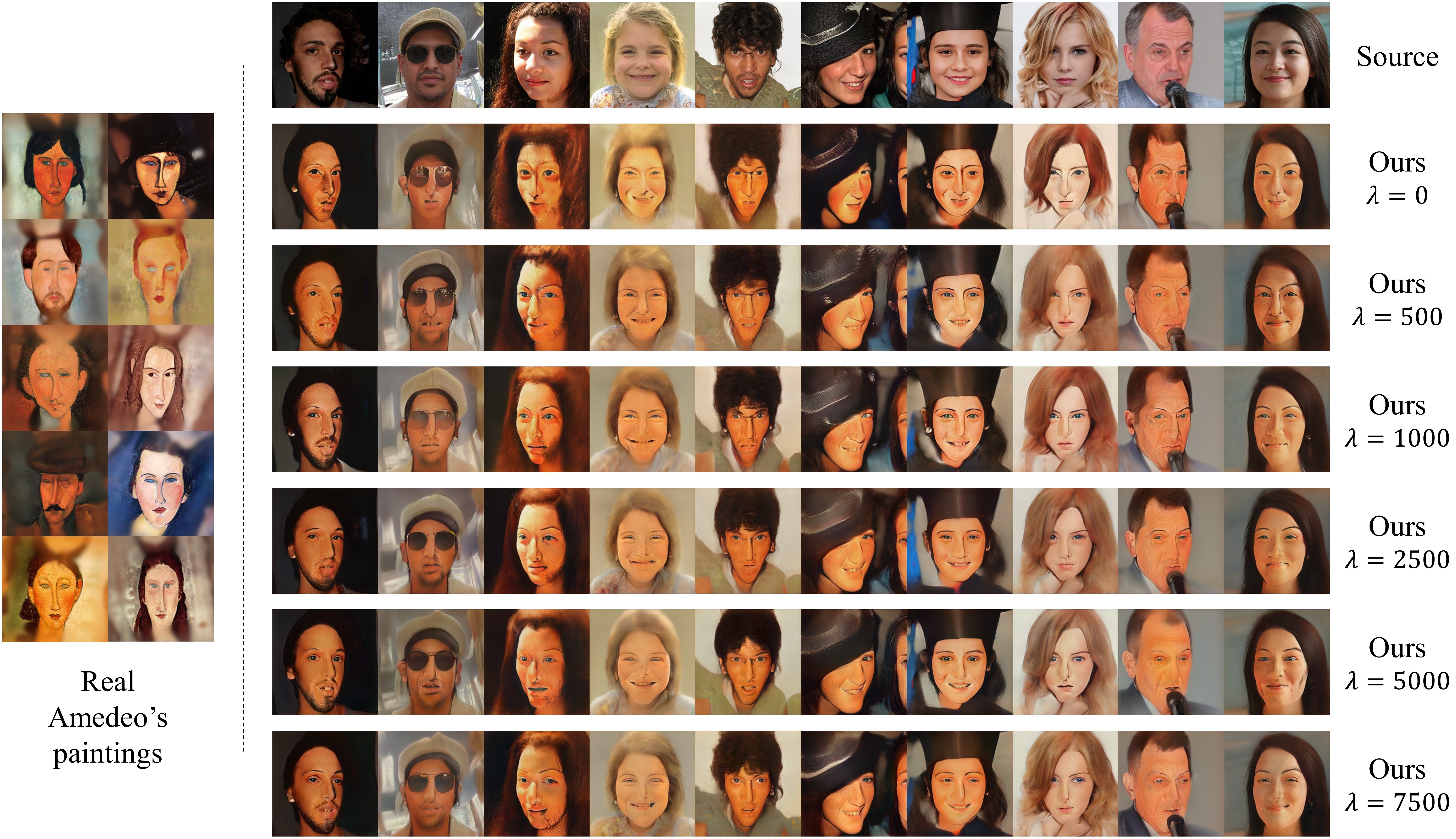}
    \caption{Visualized ablations of $\lambda$, the weight coefficient of cross-domain consistency loss applied to the generator and discriminator, on 10-shot FFHQ $\rightarrow$ Amedeo's paintings.}
    \label{lambdaablation}
\end{figure*}

\begin{figure*}[tbp]
    \centering
    \includegraphics[width=0.96\linewidth]{ 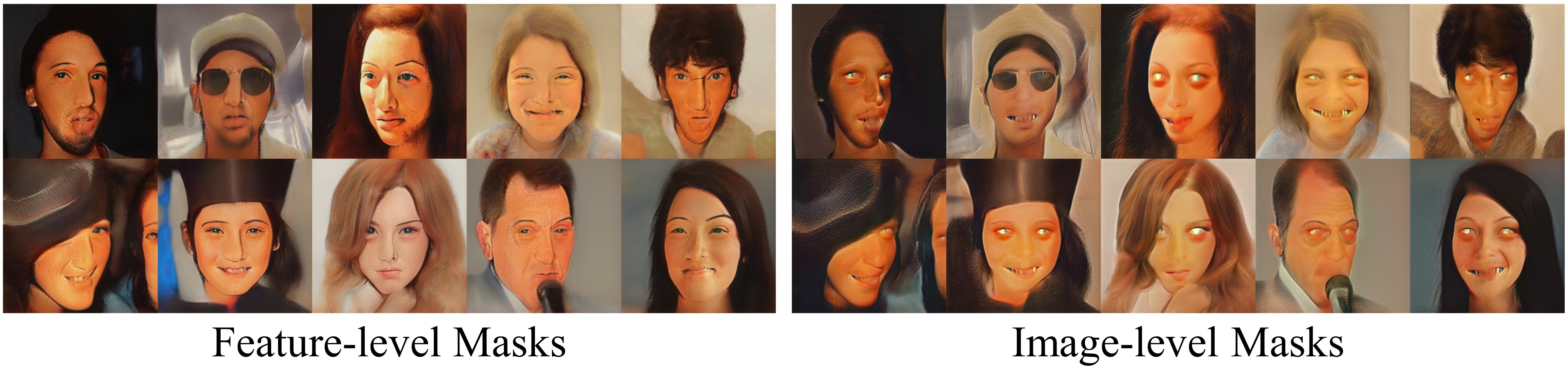}
    \caption{Visualized samples of applying image-level random masks and feature-level random masks (our approach) to the discriminator on 10-shot FFHQ $\rightarrow$ Amedeo's paintings. Samples of different models are synthesized from the same noise inputs.}
    \label{compare}
\end{figure*}

\begin{table}[t]
\centering
\setlength\tabcolsep{4.2pt}
\small
\begin{tabular}{l|c|c|c}
 \makecell[c]{Adaptation \\ Setups} & \makecell[c]{Ours w/o \\ Random Masks} & \makecell[c]{Ours w/o \\ $\mathcal{L}_{dist}(D_s,D_t)$} & Full Approach \\
\hline
\makecell[c]{FFHQ $\rightarrow$ \\ Sketches} & $0.474 \pm 0.030$ & $0.491 \pm 0.021$ & $\pmb{0.505 \pm 0.020}$ \\
\hline
\makecell[c]{FFHQ $\rightarrow$ \\ Raphael's \\paintings} & $0.569 \pm 0.016$ & $0.574 \pm 0.014$ & $\pmb{0.581 \pm 0.012}$ \\
\hline
\makecell[c]{LSUN \\ Church $\rightarrow$ \\ Haunted \\houses} & $0.625 \pm 0.013$ & $0.631 \pm 0.008$ & $\pmb{0.632 \pm 0.020}$ \\
\end{tabular}
\caption{Quantitative ablations of our approach on 10-shot FFHQ $\rightarrow$ Sketches, FFHQ $\rightarrow$ Raphael's paintings, and LSUN Church $\rightarrow$ Haunted houses in terms of Intra-LPIPS ($\uparrow$). Standard deviations are computed across 10 clusters (the same number as training samples).}
\label{qablation}
\end{table}

\begin{table}[t]
\centering
\setlength\tabcolsep{4.2pt}
\small
\begin{tabular}{l|c|c|c}
 \makecell[c]{Adaptation \\ Setups} & \makecell[c]{Ours w/o \\ Random Masks} & \makecell[c]{Ours w/o \\ $\mathcal{L}_{dist}(D_s,D_t)$} & Full Approach \\
\hline
\makecell[c]{FFHQ $\rightarrow$ \\ Sketches} & $ 38.83\pm 0.01$ & $ 34.16 \pm 0.02$ & $\pmb{28.93 \pm 0.01}$ \\
\hline
\makecell[c]{FFHQ $\rightarrow$ \\ Babies} & $ 52.91\pm 0.01$ & $ 43.63\pm 0.02$ & $\pmb{36.39 \pm 0.01}$ \\
\hline
\makecell[c]{FFHQ $\rightarrow$ \\ Sunglasses} & $ 36.89\pm 0.01$ & $ 33.22 \pm 0.03$ & $\pmb{26.96 \pm 0.01}$ \\
\end{tabular}
\caption{Quantitative ablations of our approach on 10-shot FFHQ $\rightarrow$ Sketches, FFHQ $\rightarrow$ Babies, and FFHQ $\rightarrow$ Sunglasses in terms of FID ($\downarrow$). Standard deviations are computed across 5 runs. }
\label{qablation2}
\end{table}

\textbf{Ablations of $\lambda$} $\lambda$ represents the weight coefficient of cross-domain consistency loss included in the overall loss function as shown in Equation \ref{loss}. In Fig. \ref{lambdaablation}, we visualize 10-shot image generation results on FFHQ $\rightarrow$ Amedeo's paintings using different $\lambda$ ranging from $0$ to $7500$. Unnatural blurs can be found in generated images using too small $\lambda$. With larger $\lambda$, the adapted GANs focus more on keeping characteristics learned from the source domain. Too large $\lambda$ can prevent adapted GANs from learning from the target domain and lead to unrealistic results. Here we recommend $\lambda$ ranging from $1000$ to $5000$ for adaptation setups in our paper. Different values of $\lambda$ can be tried for different adaptation setups for better results. We use the same weight coefficient for cross-domain consistency loss applied to the generator and discriminator, achieving compelling results. Combinations of different weight coefficients for the generator and discriminator can also be tried for different adaptation setups. In our experiments, we use the same weight coefficient of cross-domain consistency loss in CDC \cite{ojha2021few-shot-gan} (generators only) and our approach for fair comparison. We randomly mask 3/4 features extracted by the target discriminator in ablations of $\lambda$.

\textbf{Quantitative Ablations} We provide the quantitative ablation analysis in terms of Intra-LPIPS and FID in Table \ref{qablation} and \ref{qablation2}, respectively. Masked discrimination encourages adapted GANs to learn the common features of limited training samples, leading to better learning of target distributions and greater diversity. Discriminator CDC loss helps preserve more information learned from source domains and avoid generating blurs and artifacts, improving generation quality and diversity. Combining these two ideas helps our approach achieve pleasing visual effects and improve quantitative evaluation metrics.

\begin{table}[tbp]
\centering
\small
\begin{tabular}{l|c|c|c}
 Methods & \makecell[c]{FFHQ $\rightarrow$ \\ Babies} & \makecell[c]{FFHQ $\rightarrow$ \\ Amedeo's \\paintings} &   \makecell[c]{LSUN \\ Church $\rightarrow$ \\ Haunted \\houses} \\
\hline
\makecell[c]{Image} & $0.581 \pm 0.033$ & $0.602 \pm 0.027$ & $0.605 \pm 0.009$ \\
\hline
\makecell[c]{Feature \\ (Ours)} & $\pmb{0.595 \pm 0.006}$ & $\pmb{0.628 \pm 0.024}$ & $\pmb{0.632 \pm 0.020}$  \\
\end{tabular}
\caption{Intra-LPIPS ($\uparrow$) results comparison of applying image-level and feature-level random masks. Standard deviations are computed across 10 clusters (the same number as training samples).}
\label{comparison}
\end{table}

\begin{figure*}[tbp]
    \centering
    \includegraphics[width=1.0\linewidth]{ 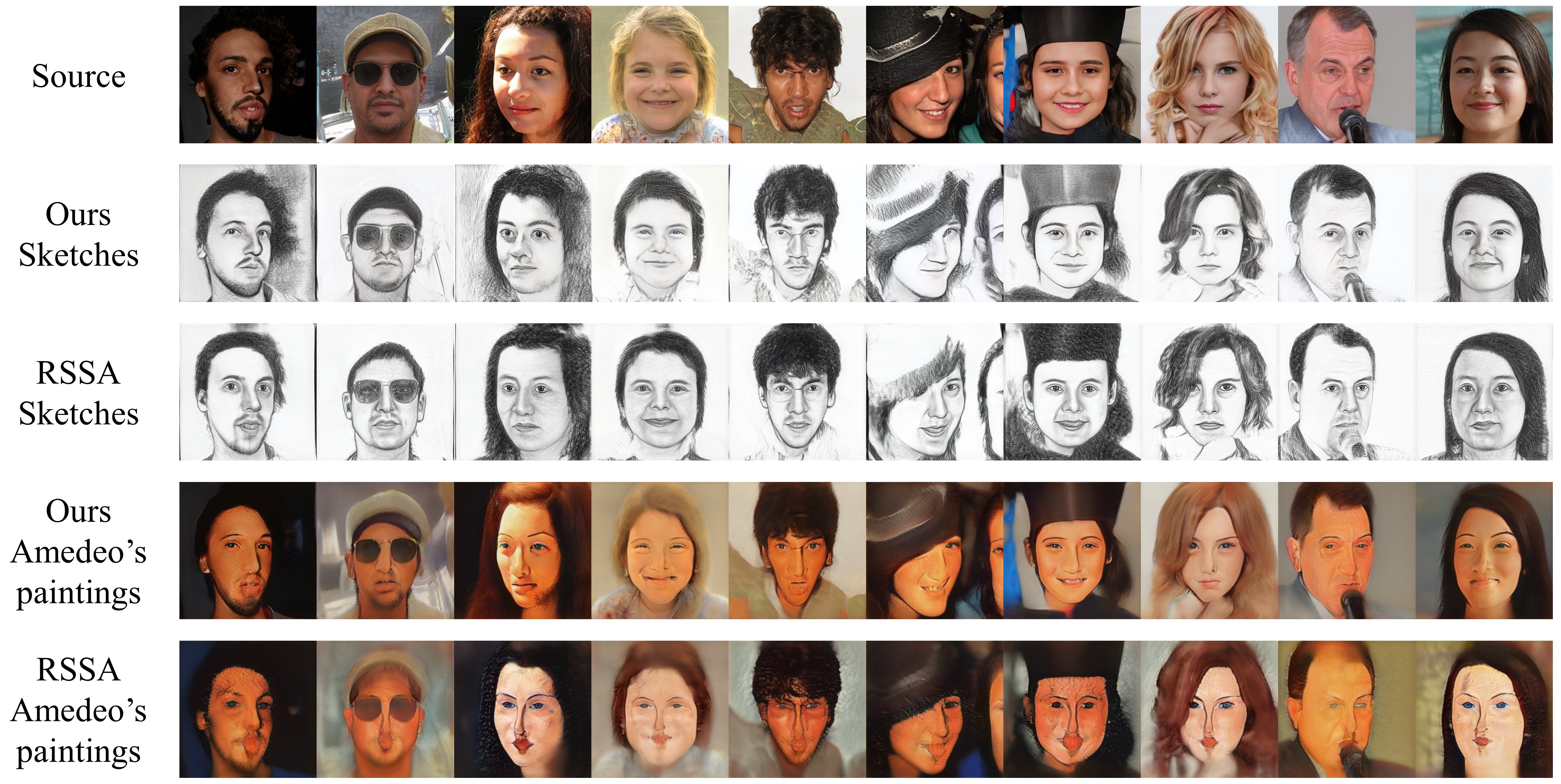}
    \caption{Visualized results comparison of our approach and RSSA \cite{xiao2022few} on 10-shot FFHQ $\rightarrow$ Sketches and FFHQ $\rightarrow$ Amedeo's paintings.  All the visualized samples of RSSA and our approach are synthesized from fixed noise inputs for fair comparison.}
    \label{RSSA}
\end{figure*}

  \begin{figure*}[tbp]
    \centering
    \includegraphics[width=1.0\linewidth]{ 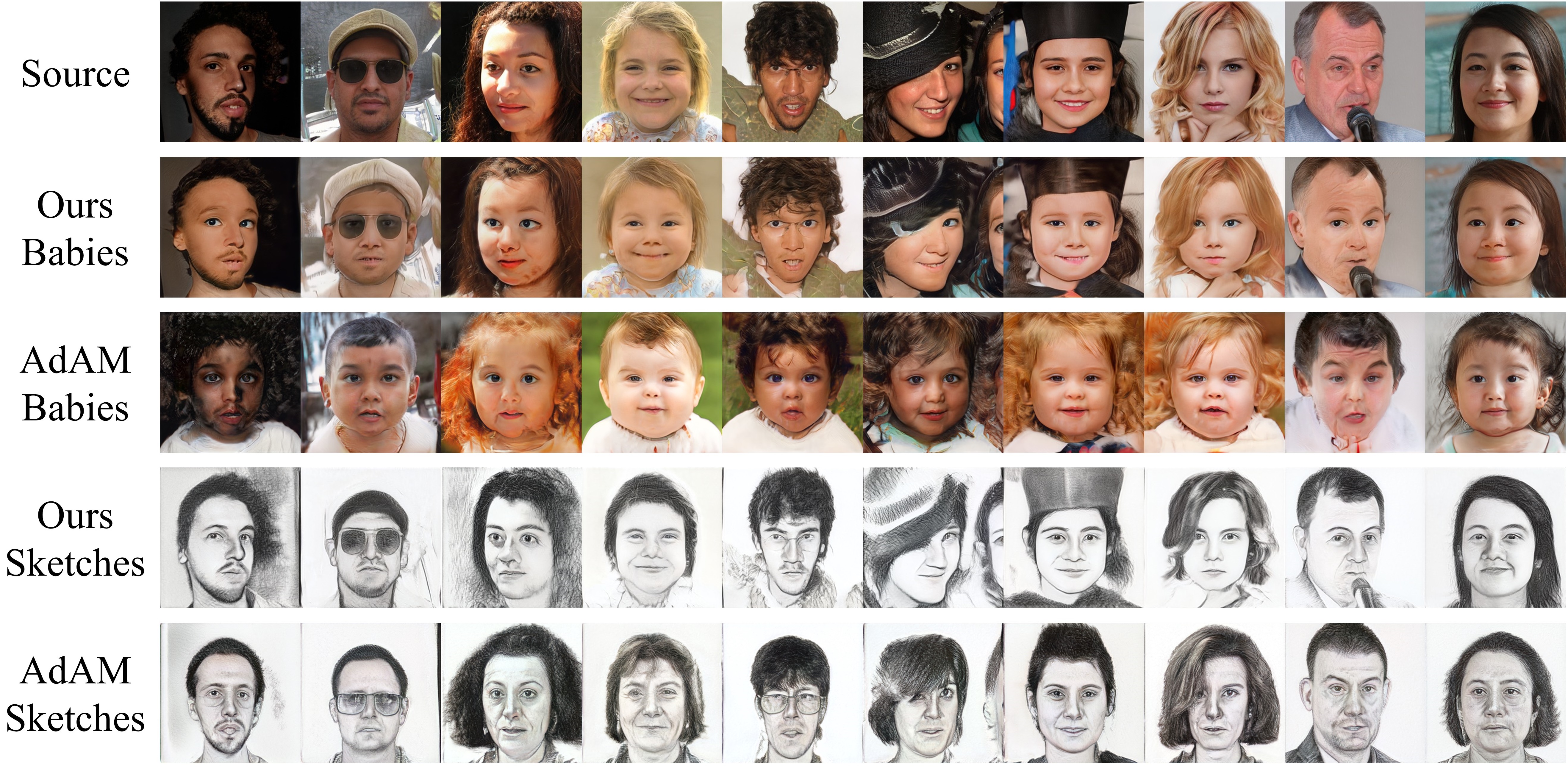}
    \caption{Visualized results comparison of our approach and AdAM \cite{adaptative} on 10-shot FFHQ $\rightarrow$ Babies and FFHQ $\rightarrow$ Sketches.  All the visualized samples of AdAM and our approach are synthesized from fixed noise inputs for fair comparison.}
    \label{adam}
\end{figure*}

\section{More Analysis of Motivation}
The proposed approach applies random masks to the output features of an intermediate layer in the discriminator. It can be seen as a feature-level augmentation approach. Here we provide comparison between applying image-level masks and our approach. Taking 10-shot FFHQ $\rightarrow$ Amedeo's paintings as an example, we show visualized samples generated from the same noise inputs in Fig. \ref{compare}. When applying random masks to input images of the discriminator, we get extremely low-quality results sharing similar structures like eyes and teeth.

In addition, image-level masks lead to lower Intra-LPIPS results as well as shown in Table \ref{comparison}. Compared with feature-level masks, masks applied to randomly chosen pixels of inputs make it harder for adapted discriminators to judge image quality. A masked high-quality image can be similar to a masked blurry image from the view of discriminators. Masked images also affect the calculation of discriminator CDC loss, leading to degraded quality and diversity.

\section{Comparison with RSSA \cite{xiao2022few}}
\label{appendixc}
RSSA proposes a relaxed spatial structural alignment method with compressed latent space based on inverted GANs \cite{Abdal_2020_CVPR}. Their approach aims to preserve the global image structures learned from source domains, which is different from our approach and other prior works listed in Table \ref{lpipsffhq}, \ref{fid}, and \ref{lsun}. Therefore we do not include RSSA as a baseline for comparison in our paper. Here we provide additional comparison between RSSA and our approach. As shown in Fig. \ref{RSSA}, RSSA focuses on global structure preservation and cannot naturally transfer images to target domains, which is inappropriate for adaptation to abstract target domains like artists' paintings. Results of RSSA are produced through the official implementation. All results are produced by adapted GANs trained for 1500 iterations in Fig. \ref{RSSA}.

\begin{figure}[t]
    \centering
    \includegraphics[width=1.0\linewidth]{ 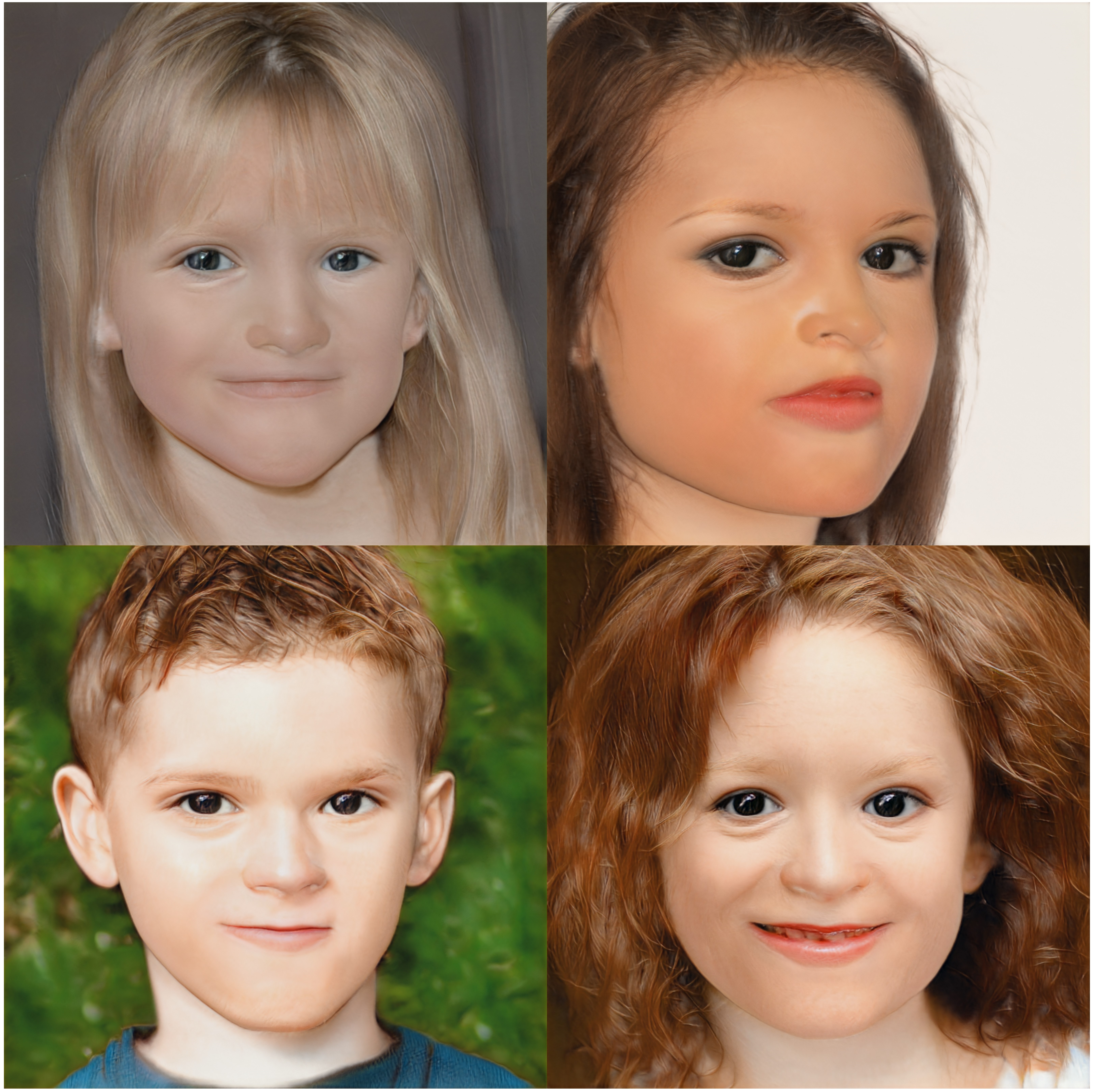}
    \caption{10-shot image generation results on $1024\times 1024$ FFHQ $\rightarrow$ Babies of our approach.}
    \label{baby1024}
\end{figure}

\section{Comparison with AdAM \cite{adaptative}}
  AdAM proposes an adaptation-aware kernel modulation for few-shot image generation, aiming to identify kernels in source GANs that are important for target adaptation. AdAM focuses on unrelated source/target domains adaptation (e.g., FFHQ $\rightarrow$ Cats, FFHQ $\rightarrow$ Cars). Their experiments are mainly conducted on target datasets containing thousands of images. Our approach is more appropriate for related source/target domains since it is designed to preserve information from source domains during domain adaptation. Besides, we focus on GAN adaptation using extremely limited training data ($\leq$ 10 images). Therefore, we do not include AdAM as a baseline for comparison in our paper. Here we add experiments of AdAM on 10-shot FFHQ $\rightarrow$ Babies and FFHQ $\rightarrow$ Sketches in comparison with our approach. As shown in Fig. \ref{adam}, our approach achieves apparently greater generation diversity and more natural adaptation for extremely limited data and related source/target domains. Visualized samples of our approach are more different from training samples and preserve more details from the source domain. Visualized samples of our approach on unrelated source/target domains are added in Appendix \ref{appendixe}. Results of AdAM are produced through the official implementation. All results are produced by adapted GANs trained for 1500 iterations in Fig. \ref{adam}.

\begin{figure}[t]
    \centering
    \includegraphics[width=1.0\linewidth]{ 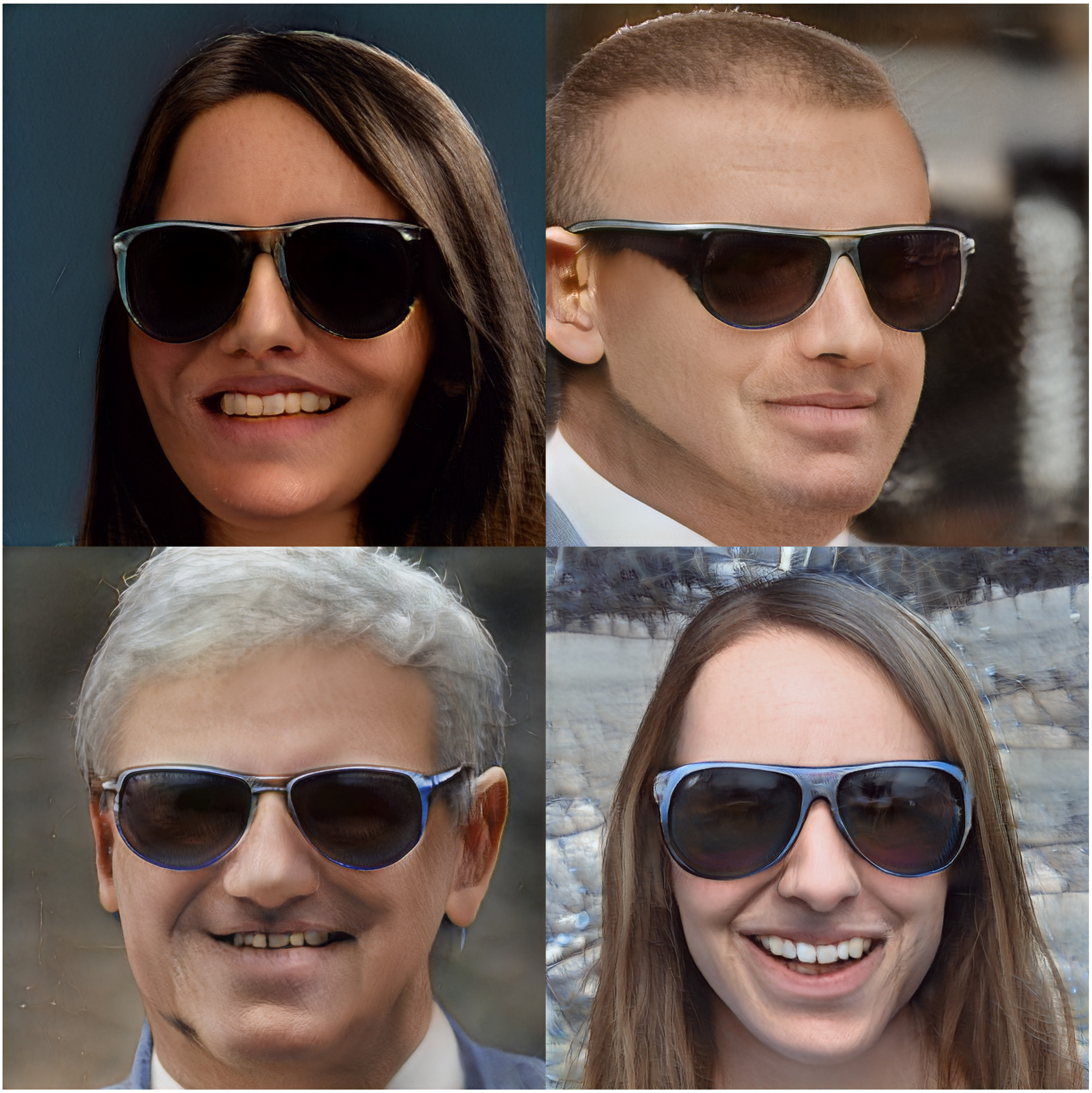}
    \caption{10-shot image generation results on $1024\times 1024$ FFHQ $\rightarrow$ Sunglasses of our approach.}
    \label{sunglass1024}
\end{figure}

\section{High-Resolution Results}
Most prior works use datasets with resolution of $128\times 128$ and $256\times 256$. In this paper, we conduct experiments with resolution of $256\times 256$, except for LSUN Cars to Wrecked cars, which has resolution of $512\times 512$. We add experiments on 10-shot FFHQ $\rightarrow$ Babies and FFHQ $\rightarrow$ Sunglasses with resolution of $1024\times 1024$. As shown in Fig. \ref{baby1024} and \ref{sunglass1024}, our approach can also be applied to high-resolution datasets and synthesize high-quality and diverse results. In addition, we compare our approach with CDC \cite{ojha2021few-shot-gan}, which performs better than other baselines on 10-shot FFHQ $\rightarrow$ Babies and Sunglasses ($256\times 256$) in terms of Intra-LPIPS in Table \ref{compare1024}. It can be seen that our approach outperforms CDC on the resolution of $1024\times 1024$ as well.

\begin{table}[t]
\centering
\begin{tabular}{l|c|c}
 Methods & \makecell[c]{FFHQ $\rightarrow$ \\ Babies} & \makecell[c]{FFHQ $\rightarrow$ \\ Sunglasses} \\
\hline
CDC \cite{ojha2021few-shot-gan} & $0.597 \pm 0.029$ & $0.606 \pm 0.017$ \\
Ours & $\pmb{0.610 \pm 0.026}$ & $\pmb{0.617 \pm 0.014}$ \\
\end{tabular}
\caption{Intra-LPIPS ($\uparrow$) results of 10-shot FFHQ $\rightarrow$ Babies and FFHQ $\rightarrow$ Sunglasses ($1024\times 1024$) using CDC and our approach. Standard deviations are computed across 10 clusters (the same number as training samples).}
\label{compare1024}
\end{table}

\begin{figure}[t]
    \centering
    \includegraphics[width=1.0\linewidth]{ 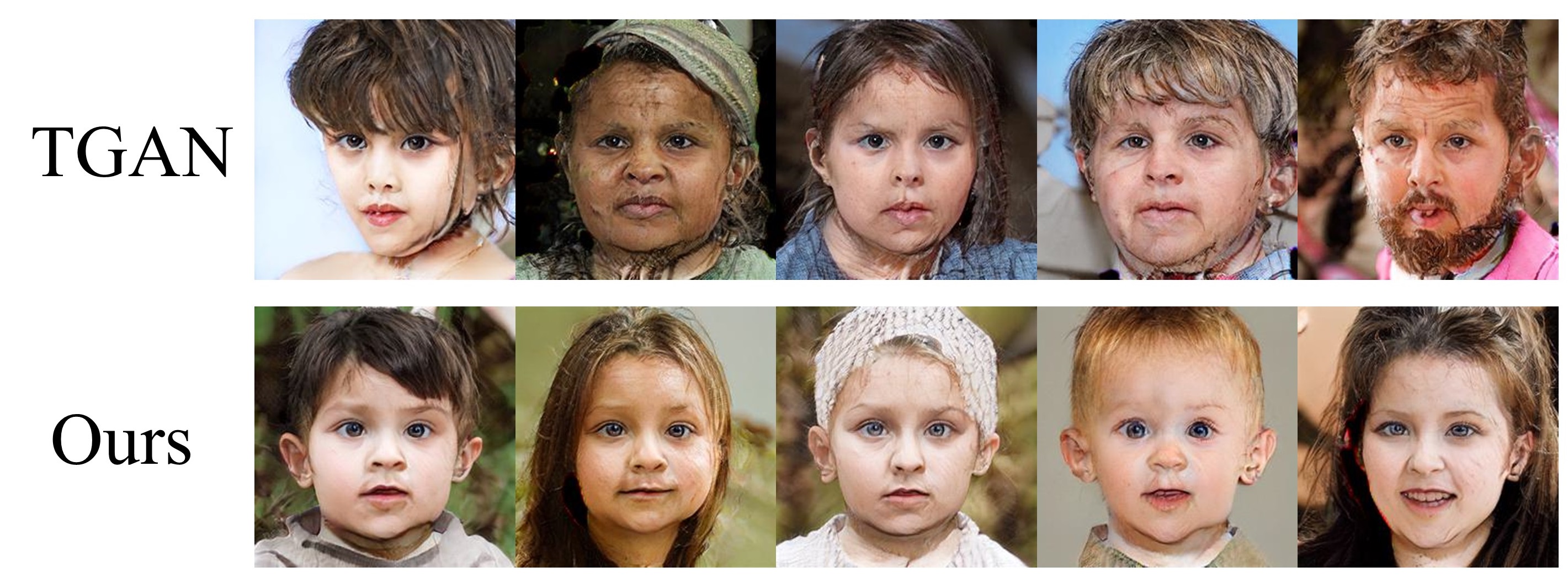}
    \caption{Visualized samples of TGAN and our approach applied to UNetGAN on 10-shot FFHQ $\rightarrow$ Babies. The results of TGAN and our approach are synthesized from the same noise inputs for fair comparison.}
    \label{unetgan}
\end{figure}

\begin{table}[t]
\centering
\setlength\tabcolsep{4pt}
\small
\begin{tabular}{l|c|c|c}
 Methods & \makecell[c]{FFHQ $\rightarrow$ \\ Sketches} & \makecell[c]{FFHQ $\rightarrow$ \\ Babies} &   \makecell[c]{FFHQ $\rightarrow$ \\ Sunglasses} \\
\hline
TGAN \cite{wang2018transferring} & $0.587 \pm 0.062$ & $0.579 \pm 0.026$ & $0.592 \pm 0.032$ \\
\hline
Ours & $\pmb{0.593 \pm 0.034}$ & $\pmb{0.590 \pm 0.023}$ & $\pmb{0.602 \pm 0.028}$  \\
\end{tabular}
\caption{Intra-LPIPS ($\uparrow$) results of our approach and TGAN applied to UNetGAN. Standard deviations are computed across 10 clusters (the same number as training samples).}
\label{biggan}
\end{table}

\section{BigGAN-based Masked Discrimination}
This paper follows prior works to apply the widely-used powerful generative model StyleGAN2 \cite{Karras_2020_CVPR} as the base model to achieve high-quality results and fair comparison. We further adapt our approach to UNetGAN \cite{schonfeld2020u}, which applies a UNet-based discriminator to BigGAN \cite{DBLP:conf/iclr/BrockDS19}. Since the cross-domain consistency loss is designed based on StyleGAN2, we directly apply masked discrimination to UNetGAN and compare our approach with the TGAN baseline. As shown in Fig. \ref{unetgan}, our approach significantly improves the generation quality with fewer blurs and preserves more details like teeth, hairstyles, and facial expressions. In addition, our approach achieves better Intra-LPIPS results than TGAN (both trained for 5000 iterations) based on the UNetGAN model, as shown in Table \ref{biggan}.

\begin{figure*}[t]
    \centering
    \includegraphics[width=1.0\linewidth]{ 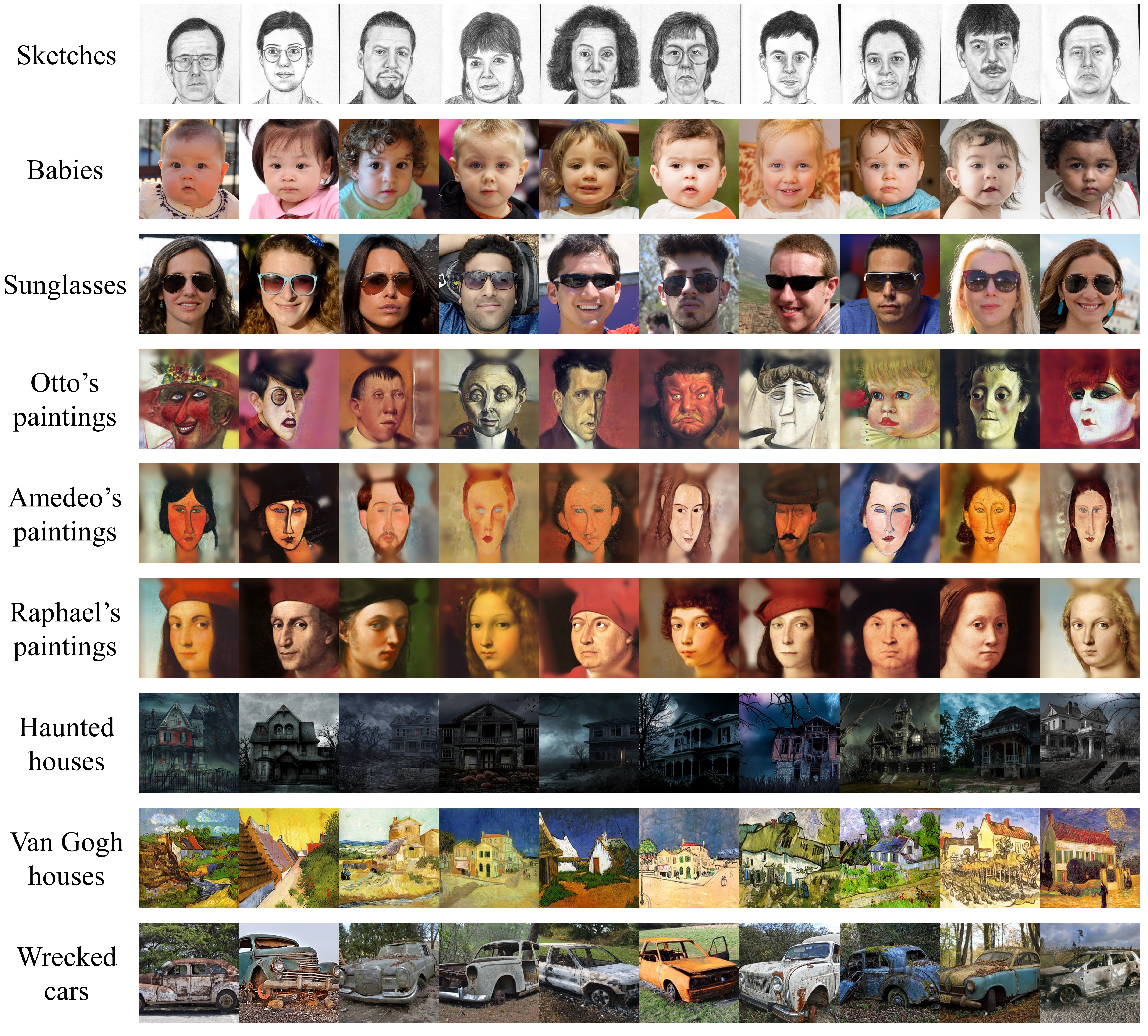}
    \caption{All the 10-shot datasets used in this paper, including 6 target domains corresponding to FFHQ, Haunted houses and Van Gogh houses corresponding to LSUN Church, and Wrecked cars corresponding to LSUN Cars.}
    \label{dataset}
\end{figure*}

\begin{figure*}[tbp]
    \centering
    \includegraphics[width=1.0\linewidth]{ 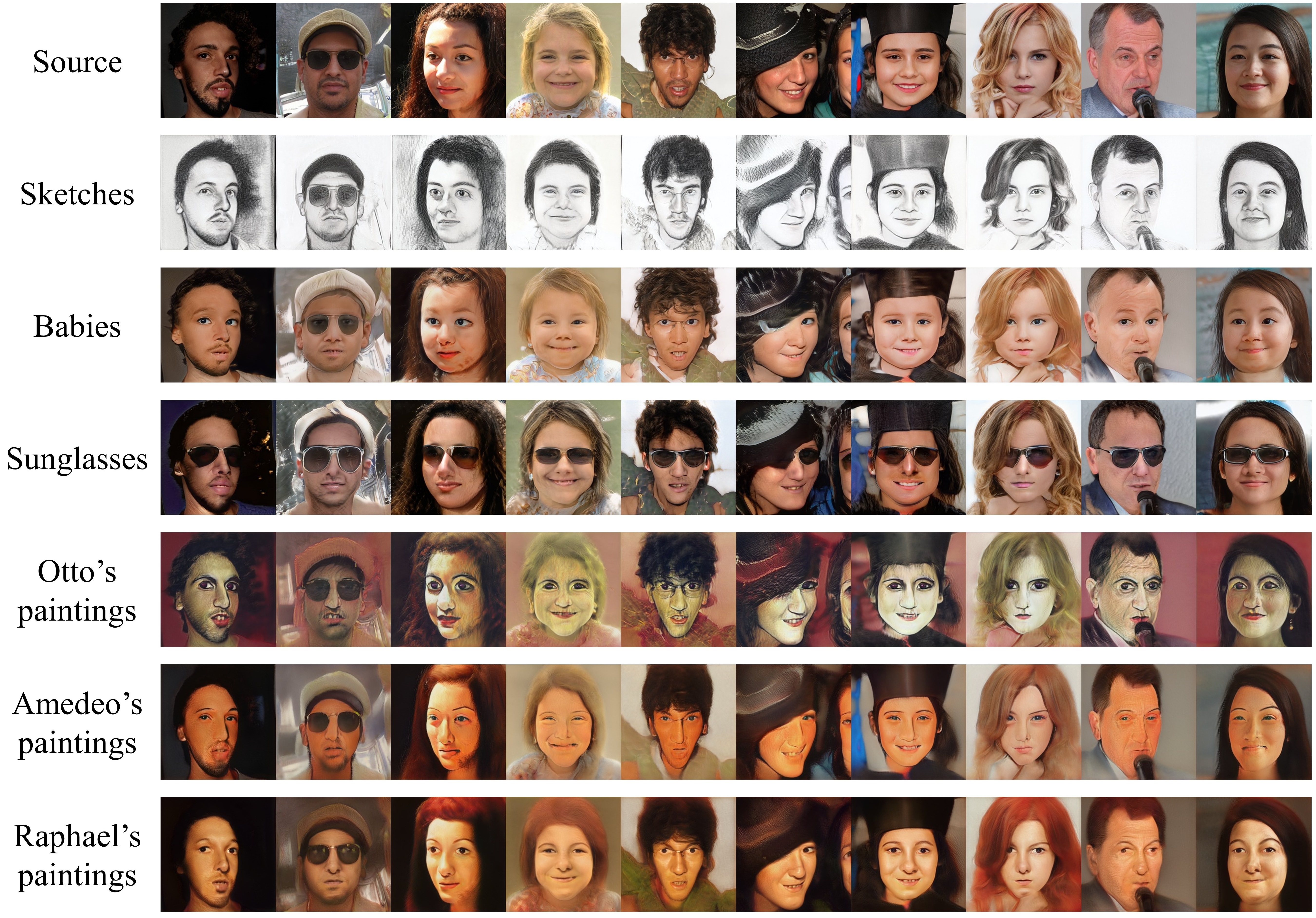}
    \caption{10-shot image generation results of our approach adapted from the source model pre-trained on FFHQ. All the visualized samples are synthesized from fixed noise inputs in adaptation to different target domains.}
    \label{ffhq_all}
\end{figure*}

\begin{table}[tbp]
\centering
\begin{tabular}{l|c}
 Methods & \makecell[c]{Time cost for 3000 iterations}  \\
\hline
TGAN \cite{wang2018transferring} & 38 min \\
TGAN+ADA \cite{ada} & 43 min \\
FreezeD \cite{mo2020freeze} & 34 min \\
MineGAN \cite{wang2020minegan} & 32 min + 38 min \\
EWC \cite{ewc} & 39 min \\
CDC \cite{ojha2021few-shot-gan} & 44 min \\
DCL \cite{zhao2022closer} &  2 h 43 min \\
RSSA \cite{xiao2022few} & 34 min + 4 h 51 min \\
AdAM \cite{adaptative} & 5 min + 59 min \\
Ours &  57 min \\
Ours ($1024 \times 1024$) & 4h \\
\end{tabular}
\caption{The time cost of 10-shot adaptation models trained for 3000 iterations on a single NVIDIA TITAN RTX GPU (image resolution $256\times256$, batch size 4).}
\label{time}
\end{table}

\section{Computational Cost}
We list the computational cost of different few-shot image generation approaches in Table \ref{time}. We conduct all the experiments under 10-shot adaptation setups (image resolution $256 \times 256)$ for 3000 iterations with the same batch size 4 on a single NVIDIA TITAN RTX GPU. MineGAN \cite{wang2020minegan} needs a two-stage training strategy. In the first stage, the generator is fixed and the miner is optimized with the discriminator. Then the miner, generator, and discriminator are optimized together in the second stage. RSSA \cite{xiao2022few} needs to project real samples into the latent space before adaptation to target domains. AdAM \cite{adaptative} needs to identify important kernels for target adaptation first. Therefore, we provide the time cost of MineGAN, RSSA, and AdAM in "X+Y" format, measuring two stages separately. The time cost of our approach is comparable to most previous methods and lower than several recent works including DCL \cite{zhao2022closer} and RSSA \cite{xiao2022few}.

\begin{figure*}[tbp]
    \centering
    \includegraphics[width=1.0\linewidth]{ 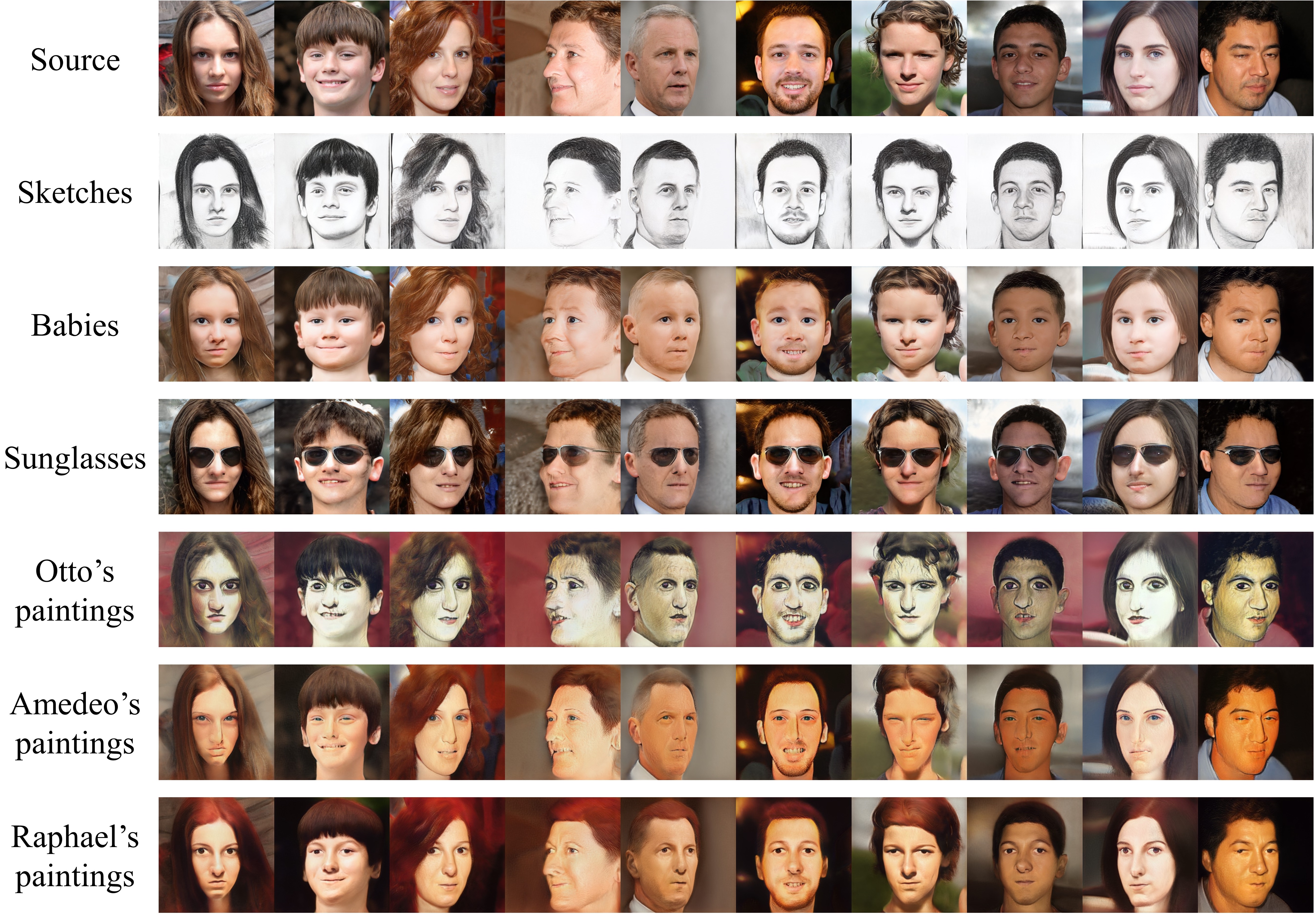}
    \caption{Additional 10-shot image generation results of our approach adapted from the source model pre-trained on FFHQ. All the visualized samples are synthesized from fixed noise inputs in adaptation to different target domains.}
    \label{ffhq_all2}
\end{figure*}

  \begin{figure*}[t]
    \centering
    \includegraphics[width=1.0\linewidth]{ 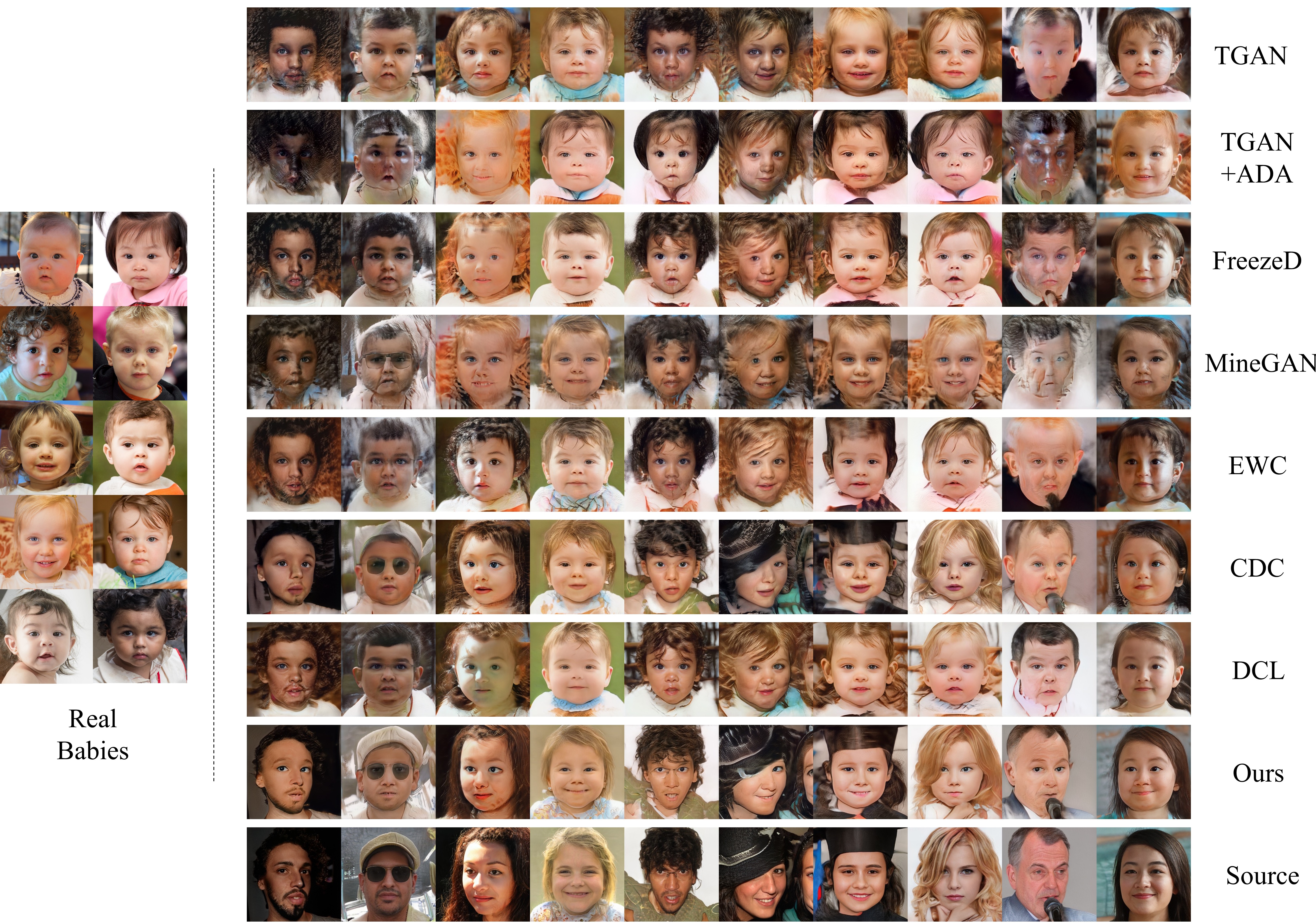}
    \caption{10-shot image generation results on FFHQ $\rightarrow$ Babies of our approach and baselines. All the visualized samples of different approaches are synthesized from fixed noise inputs for comparison.}
    \label{babies}
\end{figure*}

\section{Datasets and More Results}
\label{appendixe}

We show all the 10-shot target datasets used for the experiments in this paper in Fig. \ref{dataset}. We employ 6 target domains in correspondence to the source domain FFHQ. Fig. \ref{ffhq_all} and \ref{ffhq_all2} show few-shot adaptation examples of all these target domains generated from different noise inputs, respectively. 

\begin{figure*}[t]
    \centering
    \includegraphics[width=1.0\linewidth]{ 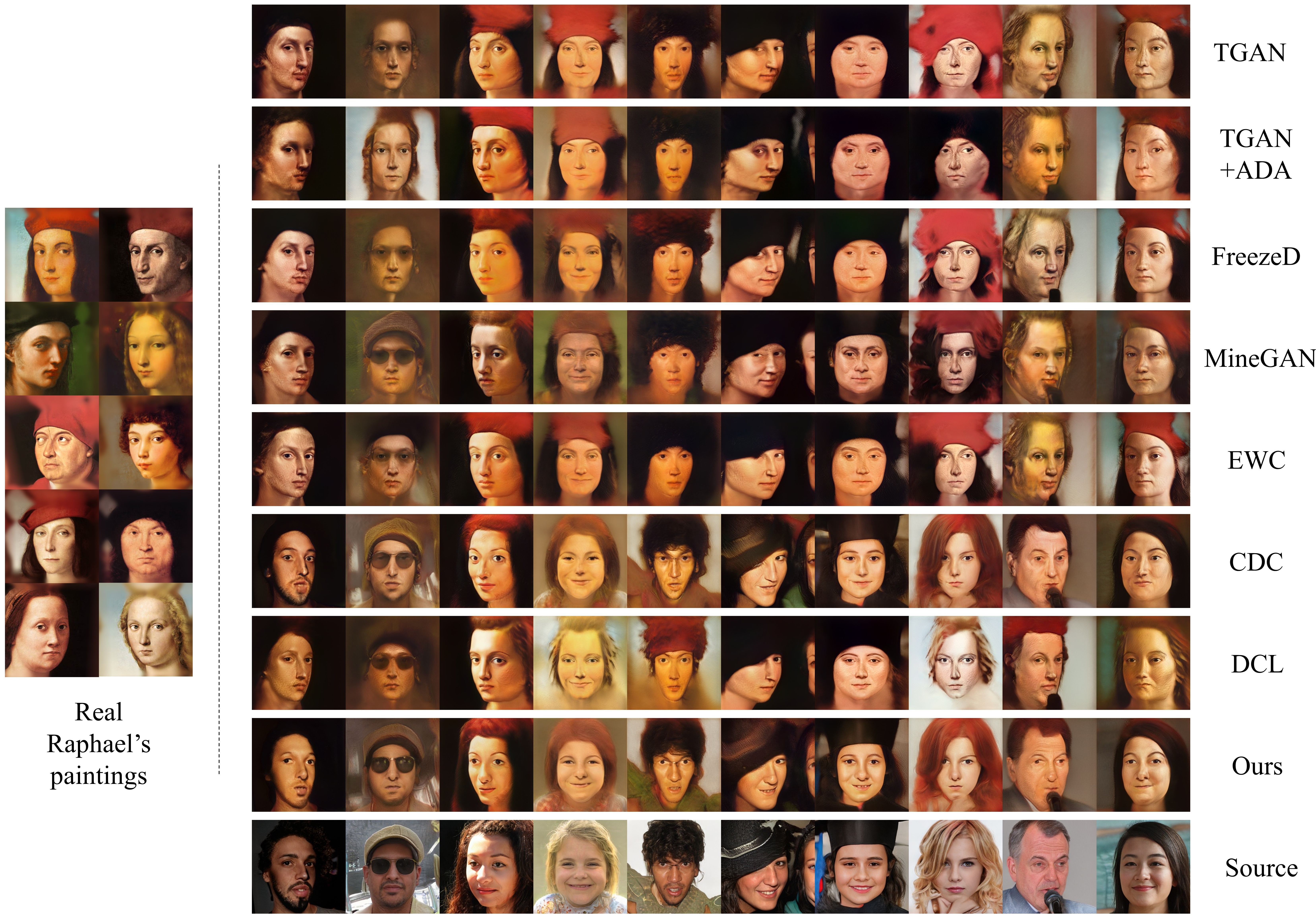}
    \caption{10-shot image generation results on FFHQ $\rightarrow$ Raphael's paintings of our approach and baselines. All the visualized samples of different approaches are synthesized from fixed noise inputs for comparison.}
    \label{raphael}
\end{figure*}

\begin{figure*}[t]
    \centering
    \includegraphics[width=1.0\linewidth]{ 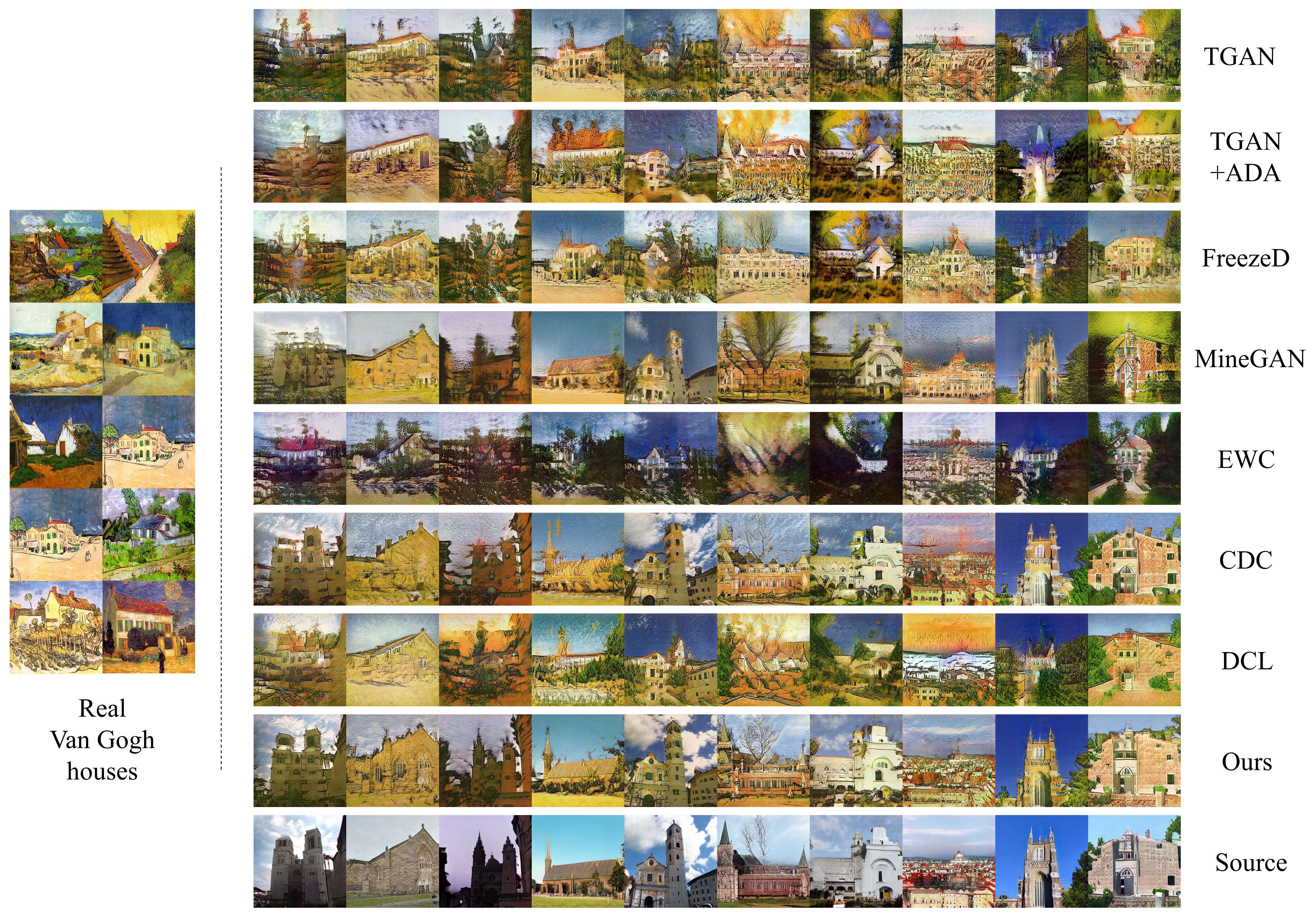}
    \caption{10-shot image generation results on LSUN Church $\rightarrow$ Van Gogh houses of our approach and baselines. All the visualized samples of different approaches are synthesized from fixed noise inputs for comparison.}
    \label{vangogh}
\end{figure*}

 In Fig. \ref{babies}, \ref{raphael}, and \ref{vangogh}, we compare our approach with baselines on 10-shot FFHQ $\rightarrow$ Babies, FFHQ $\rightarrow$ Raphael's paintings, and LSUN Church $\rightarrow$ Van Gogh houses as supplements to Fig. \ref{sketches} and \ref{amedeo}. Our approach further relieves overfitting and preserves characteristics learned from source domains better than baselines, including hairstyles, facial expressions, and house structures. Table \ref{vangoghlpips} provides additional Intra-LPIPS results on 10-shot LSUN Church $\rightarrow$ Van Gogh houses. Our approach also achieves better results than baselines.

 \begin{table}[t]
\centering
\begin{tabular}{l|c}
 Approaches & \makecell[c]{LSUN Church $\rightarrow$ \\ Van Gogh houses }  \\
\hline
TGAN \cite{wang2018transferring} & $0.611 \pm 0.016$ \\
TGAN+ADA \cite{ada} & $0.627 \pm 0.023$ \\
FreezeD \cite{mo2020freeze} & $0.606 \pm 0.018$ \\
MineGAN \cite{wang2020minegan} & $0.669 \pm 0.034$ \\
EWC \cite{ewc} & $0.611 \pm 0.029$ \\
CDC \cite{ojha2021few-shot-gan} & $0.683 \pm 0.014$ \\
DCL \cite{zhao2022closer} & $0.637 \pm 0.017$ \\
Ours & $\pmb{0.695 \pm 0.034}$ \\
\end{tabular}
\caption{Intra-LPIPS ($\uparrow$) results of 10-shot LSUN Church $\rightarrow$ Van Gogh houses. Standard deviations are computed across 10 clusters (the same number as training samples).}
\label{vangoghlpips}
\end{table}

\begin{figure*}[tbp]
    \centering
    \includegraphics[width=1.0\linewidth]{ 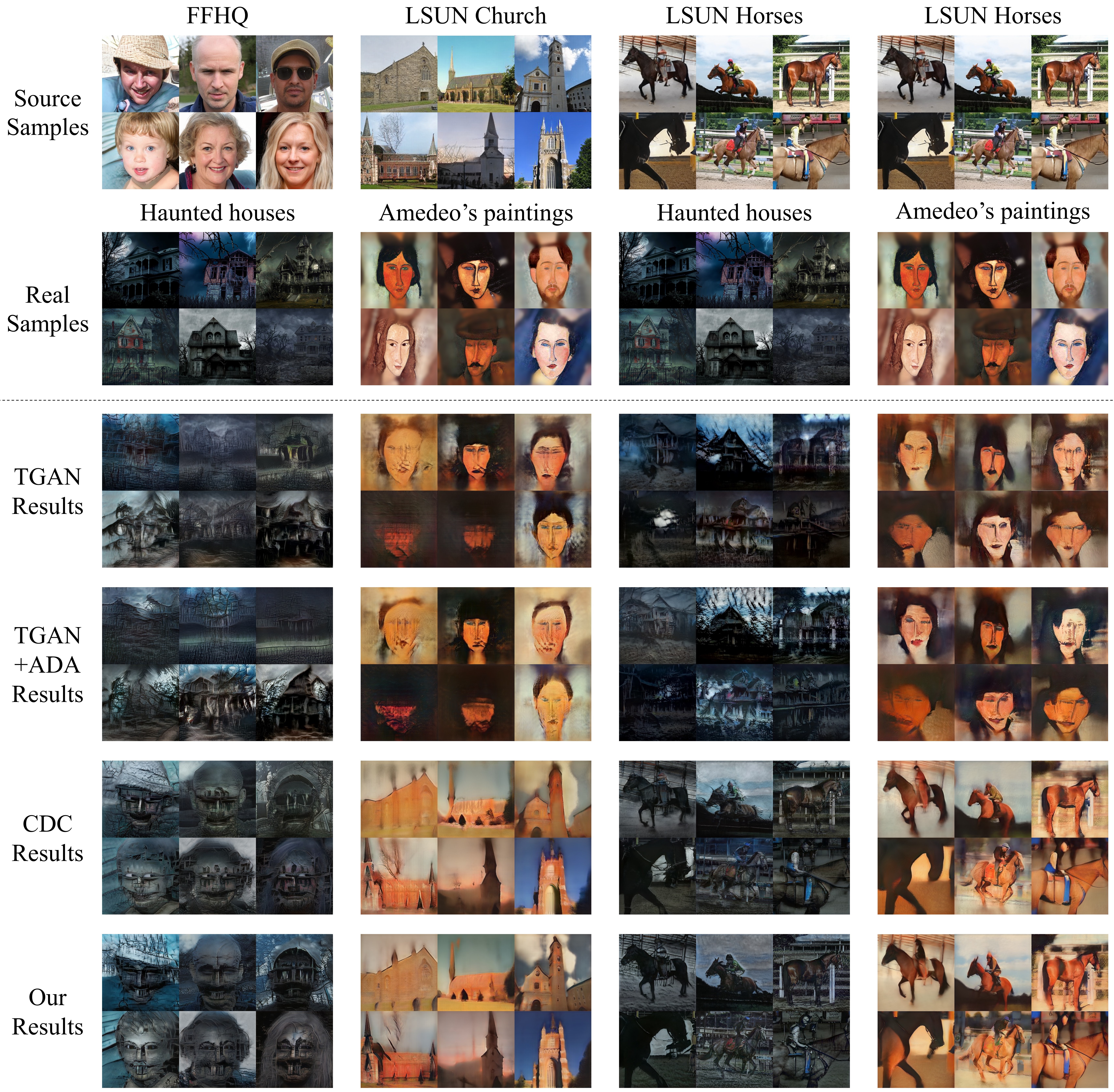}
    \caption{Visualization of 10-shot adaptation on unrelated source/target domains including FFHQ $\rightarrow$ Haunted houses, LSUN Church $\rightarrow$ Amedeo's paintings, LSUN Horses $\rightarrow$ Haunted houses, and LSUN Horses $\rightarrow$ Amedeo's paintings.}
    \label{unrelated}
\end{figure*}

Following prior works \cite{ojha2021few-shot-gan,zhao2022closer}, we add the few-shot image generation experiments on unrelated source/target domains as shown in Fig. \ref{unrelated}. TGAN \cite{wang2018transferring} and TGAN+ADA \cite{ada} tend to replicate the training samples. CDC \cite{ojha2021few-shot-gan} builds a cross-domain correspondence between the source and target domains. The proposed approach can preserve more detailed structures of faces, churches, and horses during adaptation to target domains. Our approach tends to preserve the fundamental structures of source samples instead of generating images similar to the training samples. Therefore, our approach is appropriate for related source/target domains or transferring the style of source images to target domains while keeping fundamental structures learned from source domains. Here we provide experiments on unrelated source/target domains to prove our approach's capability of preserving diverse information learned from source domains. All results are produced by adapted GANs trained for 1000 iterations in Fig. \ref{unrelated}.

\end{document}